\documentclass[lettersize,journal]{IEEEtran}
\usepackage{amsmath,amsfonts}
\usepackage{array}
\usepackage[caption=false,font=normalsize,labelfont=sf,textfont=sf]{subfig}
\usepackage{textcomp}
\usepackage{stfloats}
\usepackage{url}
\usepackage{verbatim}
\usepackage{graphicx}
\usepackage{cite}
\usepackage{bm}
\usepackage[linesnumbered,ruled,vlined]{algorithm2e}
\SetKwComment{Comment}{$\vartriangleright$ }{}

\usepackage[inline]{enumitem}
\usepackage[abbreviations]{foreign}
\graphicspath{ {figures/} }
\usepackage{tikz}
\usepackage{pgfplots}
\usepackage{sansmath}
\pgfplotsset{compat=newest}

\usepgfplotslibrary{units,colorbrewer,groupplots,fillbetween}
\usetikzlibrary{patterns}

\makeatletter
\begingroup\endlinechar=-1\relax
\everyeof{\noexpand}%
\edef\x{\endgroup\def\noexpand\homepath{%
		\@@input|"kpsewhich --var-value=HOME" }}\x
\makeatother

\usepackage{acronym}
\usepackage{booktabs}
\usepackage{xspace}
\usepackage{xfrac}
\usepackage[per-mode=symbol, detect-weight=true, detect-family=true]{siunitx}
\usepackage{multicol}
\usepackage{multirow}
\usepackage{amsmath, amsfonts, amssymb}
\usepackage{amsthm}
{
	\theoremstyle{plain}
	\newtheorem{assumption}{Assumption}
}
{
	\theoremstyle{plain}
	\newtheorem{theorem}{Theorem}
}
{
	\theoremstyle{plain}
	\newtheorem{lemma}[theorem]{Lemma}
}

\usepackage[hidelinks]{hyperref}
\usepackage{cleveref}

\acrodef{DL}{deep learning}
\acrodef{ML}{machine learning}
\acrodef{FL}{federated learning}
\acrodef{SGD}{stochastic gradient descent}
\acrodef{IID}{independent and identically distributed}
\acrodef{non-IID}{non independent and identically distributed}
\acrodef{HPO}{hyperparameter optimization}
\acrodef{HP}{hyperparameter}
\acrodef{ClientOpt}{client-side optimizer}
\acrodef{ServerOpt}{server-side optimizer}
\acrodef{SOTA}{state-of-the-art}
\acrodef{DNN}{deep neural network}

\newcommand{\sys}{\textsc{GeL}\xspace}
\newcommand{\gel}{\textsc{GeL}\xspace}
\newcommand{\fednova}{\textsc{FedNova}\xspace}
\newcommand{\fedavg}{\textsc{FedAvg}\xspace}

\newcommand{\fedavgCM}{\textsc{FedAvgCM}\xspace}

\newcommand{\fedprox}{\textsc{FedProx}\xspace}

\newcommand{\fedadam}{\textsc{FedAdam}\xspace}
\newcommand{\fedyogi}{\textsc{FedYogi}\xspace}

\newcommand{\adam}{\textsc{Adam}\xspace}
\newcommand{\yogi}{\textsc{Yogi}\xspace}
\newcommand{\sgd}{\textsc{SGD}\xspace}
\newcommand{\sgdm}{\textsc{SGDM}\xspace}
\newcommand{\aqfl}{\textsc{AQFL}\xspace}
\newcommand{\fedopt}{\textsc{FedOpt}\xspace}
\newcommand{\copt}{\textsc{ClientOpt}\xspace}
\newcommand{\sopt}{\textsc{ServerOpt}\xspace}
\newcommand{\flora}{\textsc{FLoRA}\xspace}
\newcommand{\heteroFL}{\textsc{HeteroFL}\xspace}
\newcommand{\adaptCL}{\textsc{AdaptCL}\xspace}
\newcommand{\scalefl}{\textsc{ScaleFL}\xspace}
\newcommand{\FL}{\ac{FL}\xspace}

\newcommand{\fedavgCMG}{\textsc{FedAvgCM + GeL}\xspace}

\newcommand{\fedyogiG}{\textsc{FedYogi + GeL}\xspace}

\newcommand{\scaleflg}{\textsc{ScaleFL + GeL}\xspace}

\newcommand{\obj}{F}
\newcommand{\bnda}{\beta^2}
\newcommand{\bndb}{\kappa^2}
\newcommand{\tg}{\nabla F}
\newcommand{\nworkers}{m}
\newcommand{\lip}{L}
\newcommand{\vbnd}{\sigma^2}
\newcommand{\surloss}{\smash{\widetilde{F}}}

\newcommand{\cpavg}{\overline{\tau}}
\newcommand{\cp}{\tau}
\newcommand{\cpp}{\tau'}
\newcommand{\cpeff}{\tau_{\text{eff}}}
\newcommand{\teff}{\tau_{\text{eff}}}
\newcommand{\x}{\boldsymbol{x}\xspace}

\newcommand{\thlf}{\sfrac{3}{2}}

\newcommand{\vsym}{\boldsymbol{v}\xspace}
\newcommand{\dsym}{\boldsymbol{d}\xspace}
\newcommand{\asym}{\boldsymbol{a}\xspace}
\newcommand{\gsym}{\boldsymbol{g}\xspace}
\newcommand{\Gsym}{\boldsymbol{G}\xspace}
\newcommand{\xsym}{\boldsymbol{x}\xspace}

\newcommand{\ysym}{\hat{y}\xspace}

\newcommand{\clr}{\eta_l}
\newcommand{\slr}{\eta}

\newcommand{\cifar}{CIFAR-10\xspace}
\newcommand{\cifarH}{CIFAR-100\xspace}
\newcommand{\agnews}{AG-News\xspace}

\newcommand\copyrighttext{%
    \footnotesize \textcopyright 2025 IEEE.
    Personal use of this material is permitted.
    Permission from IEEE must be obtained for all other uses, in any current or future media, including reprinting/republishing this material for advertising or promotional purposes, creating new collective
    works, for resale or redistribution to servers or lists, or reuse of any copyrighted component of this work in other works.
    Pre-print version. Published in the IEEE Transactions on Parallel and Distributed Systems.
    For the final published version, please refer to DOI \href{https://doi.org/10.1109/TPDS.2025.3578522}{10.1109/TPDS.2025.3578522}.}
\newcommand\copyrightnotice{%
    \begin{tikzpicture}[remember picture,overlay]
        \node[anchor=south,yshift=0pt,fill=yellow!20] at (current page.south) {\fbox{\parbox{\dimexpr\textwidth-\fboxsep-\fboxrule\relax}{\copyrighttext}}};
    \end{tikzpicture}%
}

\begin{document}
\title{Boosting Resource-Constrained Federated Learning Systems with Guessed Updates}

\author{Mohamed~Yassine~Boukhari,
	Akash~Dhasade,
	Anne-Marie~Kermarrec,\\
	Rafael~Pires,
	Othmane~Safsafi, and
	Rishi~Sharma%
	\thanks{Mohamed~Yassine~Boukhari, 	Akash~Dhasade, Anne-Marie~Kermarrec, Rafael~Pires, and Rishi~Sharma are with EPFL, Switzerland}%
	\thanks{Othmane~Safsafi is an independent researcher}%
	\thanks{Corresponding author: Akash~Dhasade}%
	\thanks{Email: akash.dhasade@epfl.ch}
}

\markboth{IEEE Transactions on Parallel and Distributed Systems}%
{Shell \MakeLowercase{\textit{et al.}}: A Sample Article Using IEEEtran.cls for IEEE Journals}

\maketitle
\copyrightnotice
\pagestyle{plain}

\begin{abstract}
\Ac{FL} enables a set of client devices to collaboratively train a model without sharing raw data.
This process, though, operates under the constrained computation and communication resources of edge devices.
These constraints combined with systems heterogeneity force some participating clients to perform fewer local updates than expected by the server, thus slowing down convergence.
Exhaustive tuning of hyperparameters in \Ac{FL}, furthermore, can be resource-intensive, without which the convergence is adversely affected.
In this work, we propose \sys, the guess and learn algorithm.
\sys enables constrained edge devices to perform additional learning through guessed updates on top of gradient-based steps.
These guesses are \emph{gradientless}, \ie, participating clients leverage them \emph{for free}.
Our generic guessing algorithm
\begin{enumerate*}[label=\emph{(\roman*)}]
	\item can be flexibly combined with several state-of-the-art algorithms including \fedprox, \fednova, \fedyogi or \scalefl; and
	\item achieves significantly improved performance when the learning rates are not best tuned.
\end{enumerate*}
We conduct extensive experiments and show that \sys can boost empirical convergence by up to 40\% in resource-constrained networks while relieving the need for exhaustive learning rate tuning.
\end{abstract}


\section{Introduction}
\label{sec:introduction}

In recent years, the proliferation of Internet of Things (IoT) devices, smartphones, and other edge devices has led to an unprecedented growth in data generated at the edge~\cite{NAYAK2024783}.
This data is often rich in context and diversity, making it highly valuable for training machine learning (ML) models and improving their quality~\cite{10234250}.
However, privacy concerns and regulatory constraints have made centralized training approaches increasingly impractical, as they require transferring sensitive raw data to a central server.
This has driven the development of decentralized learning paradigms that respect user privacy while leveraging the computational and data resources at the edge.

Federated Learning (FL)~\cite{mcmahan2017communication} has become a compelling technique for training \ac{ML} models in a network of remote devices.
\FL allows participating nodes to collaboratively train a single model without sharing raw data, thus ensuring a certain level of privacy while exploiting edge resources.
This paradigm has recently received considerable attention from academia and industry~\cite{ULLAH2023423,AKHTARSHENAS2024107964,yang2021characterizing,MLSYS2019_bd686fd6,yang2018applied}.

More specifically, in \ac{FL}, a central server initiates a training round by broadcasting a global model to participating client devices.
The server requests the clients to train for a fixed number of steps and waits for a stipulated amount of time to receive the locally trained models~\cite{mcmahan2017communication,MLSYS2019_bd686fd6}.
These models are then aggregated to compose the new global model to be iteratively trained again by a new set of clients in the next round.
Throughout this iterative process, the central server is responsible solely for aggregation and orchestration, while all model updates are computed locally by the client devices.
The training continues in this manner until a predefined stopping criterion is met.

\ac{FL} has developed rapidly, yet it has difficulties when training models at the edge.
Clients at the edge are heterogeneous, both in compute and communication capabilities.
The local model training and its transfers with the server can be resource-intensive for clients.
For instance, slow clients are often discarded from training for not meeting the deadlines due to poor network connection, lack of memory, or slow processing~\cite{10.1145/3552326.3567485,MLSYS2019_bd686fd6,kairouz2019advances}.
In the same time frame, faster clients are able to perform more local updates within the stipulated time window.
Such systems heterogeneity results in slower convergence which can make \FL training excessively slow, with training tasks possibly taking up to a few days~\cite{MLSYS2019_bd686fd6}.

We refer to the number of local model updates that a client $i$ can perform in a given round as its \textit{computational budget}, denoted by $\cp_i$.
The server's expected number of update steps within a stipulated time window is denoted by $\cp$.
Due to the stringent resource constraints of client devices, $\cp_i$ is typically less than $\cp$.
Furthermore, $\cp_i$ often varies across clients, a phenomenon referred to as \textit{device } or \textit{compute heterogeneity}~\cite{10.1145/3552326.3567485,abdelmoniem:2021:heterogeneity}.
Existing research has tackled this challenge by developing optimization techniques to improve convergence under heterogeneous budgets~\cite{li2018federated,Karimireddy2020SCAFFOLDSC} and by devising aggregation rules that effectively combine updates from clients with varying computational capabilities~\cite{QI2024272,NEURIPS2020_564127c0}.
Other approaches attempt to mitigate compute heterogeneity by over-allocating clients to ensure that enough complete the expected number of steps~\cite{MLSYS2019_bd686fd6}.
However, this strategy wastes resources by discarding unused models and may introduce bias by favoring faster clients~\cite{MLSYS2022_f340f1b1}.

\begin{figure}[t]
	\centering
	\includegraphics{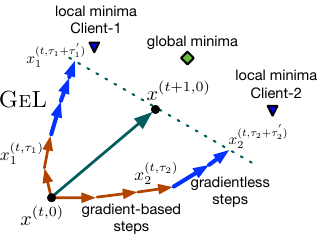}
	\caption{
				Guessing in \sys:
				The server broadcasts $\xsym^{(t, 0)}$ to participating clients for local training in the $t$-th communication round.
				Clients $1$ and $2$ first perform gradient based updates to arrive at $\xsym_1^{(t, \cp_1)}$ and $\xsym_2^{(t, \cp_2)}$.
				Next, they perform gradientless guessed updates using \sys, to arrive at $\xsym_1^{(t, \cp_1 + \cp'_1)}$ and $\xsym_2^{(t, \cp_2 + \cp'_2)}$.
				These guessing steps effectively speed up the convergence of the aggregate model $\xsym^{(t+1, 0)}=$\texttt{ Agg}$(\xsym_1^{(t, \cp_1 + \cp'_1)},\xsym_2^{(t, \cp_2 + \cp'_2)})$ towards the global minima.
	}
	\label{fig:GeL_explanation}
\end{figure}

In this work, we take an orthogonal approach and design \sys, a novel \emph{guess and learn} algorithm.
\sys enables constrained clients at the network edge to perform additional local learning through \textit{guessed} updates, compensating for undone work ($\cp - \cp_i$).
Instead of $\cp_i$ local updates, clients in \sys perform $(\cp_i + \cpp_i)$ updates, with $\cp_i$ gradient-based updates and $\cpp_i$ guessed updates.
The power of \sys lies in the fact that these guesses come for \emph{free}, \ie without entailing any extra gradient computations.
By virtually achieving the expected steps $\cp$, \sys alleviates the impact of systems heterogeneity and boosts convergence.
 \Cref{fig:GeL_explanation} illustrates how \sys operates.

\sys leverages local momentum at clients to perform guessed updates.
In \ac{ML}, momentum is a technique that accelerates convergence by maintaining a running average of past gradients, which helps navigate flat or oscillatory regions in the optimization landscape~\cite{pmlr-v28-sutskever13}.
When clients train for $\cp_i$ local steps, they accumulate local momentum, encoding valuable information about the direction and magnitude of the next update.
\sys uses this information to guide clients in taking additional steps in the momentum's direction, producing informed guessed updates.
Our generic guessing procedure features two advantages.
First, it can be flexibly applied on top of several \FL algorithms, including \fedprox~\cite{li2018federated}, \fednova~\cite{NEURIPS2020_564127c0}, \fedyogi~\cite{reddi2021adaptive} and \scalefl~\cite{ilhan2023scalefl}.
Such combinations leverage \sys (for fast empirical convergence) while tackling specific problems (\eg, resource adaptiveness in \scalefl) in the face of constrained computational budgets.
Second, it reduces the need for extensive learning rate tuning.

The guessing procedure in \sys significantly improves the performance of bad learning rate parameters, which otherwise would achieve deteriorated test performance.
A crucial component of any \ac{ML} training pipeline is \ac{HPO}~\cite{lavesson:tuningImpact,mantovani:toTune,probst2018tunability,weerts2020importance}.
Among several hyperparameters, tuning learning rate is known to be particularly important~\cite{keskar2017largebatch,nar2018stepsize,NEURIPS2018_lr,charles2020outsized}.
However, when training on decentralized data, this tuning of \acp{HP} is notoriously expensive and time-consuming~\cite{kairouz2019advances}.
Adaptive methods like \fedyogi~\cite{reddi2021adaptive} increase the range of well-performing parameter values but still require grid inspection of different client and server learning rates $(\clr, \slr)$.
Our empirical results show that \sys can enhance the test performance across a large set of values in this grid, thus significantly relieving the need for expensive tuning of learning rates.

\textbf{Contributions.}
In the light of the above, we make the following main contributions.
\begin{itemize}
	\item We introduce \sys, a novel algorithm that compensates for limitations of constrained devices and varying system capabilities in \ac{FL} by enabling guessed model updates for free (\Cref{subsec:gel_main}).
	\item  Based on the general theoretical framework of Wang \etal~\cite{NEURIPS2020_564127c0}, we provide a convergence analysis along with insights into the coefficients of gradient combinations induced by \sys that lead to improved performance (\Cref{subsec:convergence_analysis,subsec:gel_insights}).
	\item We conduct extensive experiments on $6$ different learning tasks and demonstrate that \sys converges up to 30\% faster in the number of communication rounds than the \fedavg baseline with client momentum (\Cref{subsec:gel_vs_baseline_eval,subsec:additional_datasets}).
	\item We show that \sys  can be flexibly combined with state-of-the-art FL algorithms while boosting their empirical performance by up to 40\% (\Cref{subsec:gel_against_sota_eval,subsec:eval_gel_in_fedopt,subsec:gel_vs_scalefl}).
	\item We demonstrate the benefits of \sys in improving test accuracy across a range of learning rate values. Our results show that using \sys eases tuning while achieving fast empirical convergence (\Cref{subsec:gel_untuned_eval,subsec:eval_gel_in_fedopt}).

\end{itemize}

\section{Related work and background}
\label{sec:rw_and_background}

We start by describing the \ac{FL} system in \Cref{subsec:background_FL}.
\Cref{subsec:background_fedavg_alg} describes the \fedavg algorithm along with a more formal introduction of the \ac{FL} setup.
Then we discuss a few algorithms designed to address compute or systems heterogeneity in \Cref{subsec:algorithms_for_htr}.
\Cref{subsec:hp_in_FL} presents an overview of hyperparameter optimization in \FL.
\Cref{subsec:utility_of_mom} provides an overview of client local momentum and demonstrates its utility in speeding up \FL convergence which leads to the design of \sys in \Cref{sec:gel_algo}.

\subsection{The Federated Learning (FL) system}
\label{subsec:background_FL}
\ac{FL} operates in a decentralized setting where a central server coordinates a group of distributed devices (\eg smartphones or IoT devices) to collaboratively train a shared global model.
In this paradigm, introduced with the Federated Averaging (\fedavg) algorithm~\cite{mcmahan2017communication}, the devices, referred to as clients, maintain private local datasets that remain on-device throughout the training process.
As the data never leaves the devices, \ac{FL}
enables privacy-preserving
collaborative training of the shared model.

\begin{figure}[t]
	\centering
	\includegraphics{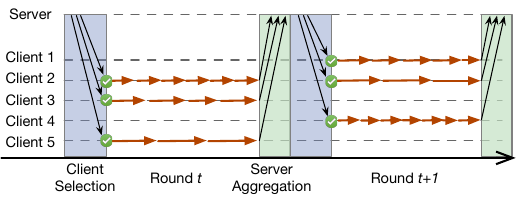}
	\caption{
		Progression of \ac{FL} timeline.
		Participating clients (\textcircled{$\checkmark$}) in each round, receive the global model from the server for local training.
		Depending upon their respective system constraints, each client completes a different number of local training steps within the stipulated time window.
		The clients then return the updated model to the server for aggregation as the process repeats.
	}
	\label{fig:GeL_timeline}
\end{figure}

A schematic of the \ac{FL} workflow across time is shown in \Cref{fig:GeL_timeline}.
The training proceeds over a series of communication rounds.
At the start of each round, the server selects a subset of clients -- designated as participants -- for the current round from the pool of available clients.
These participants receive the latest global model and associated configuration parameters, such as hyperparameter settings, from the server.
Each participant then performs local training on its private data for a specified number of steps and computes model updates -- representing the changes relative to the global model -- which are sent back to the server.
The server then aggregates them (e.g., using weighted averaging) to produce the new global model.
This process repeats over multiple rounds until a predefined stopping criterion is met, such as achieving a target accuracy.

To maintain the efficiency of this process, the server typically enforces a reporting deadline, at which point the clients must send back the updated model.
Due to the differences in the system characteristics, such as the compute speed, network bandwidth, memory, \etc, not every client is able to complete the target number of steps by the reporting deadline.
This compute heterogeneity affects the convergence of \ac{FL} algorithms, with training tasks possibly taking many communication rounds spanning over several days~\cite{MLSYS2019_bd686fd6}.
The goal of this work is to boost the convergence of \ac{FL} algorithms operating in such resource-constrained networks of clients.

\subsection{The \fedavg algorithm}
\label{subsec:background_fedavg_alg}

\textbf{Setup.} We consider a system with $m$ total clients.
We refer to the global model with the server in the $t$-th communication round as $\xsym^{(t, 0)}$.
The local model at the $i$-th client after $k$ training steps in the $t$-th communication round is referred to as $\xsym_i^{(t, k)}$.
The server selects a subset of clients $S^{(t)} \subseteq \{1,2,\ldots,m\}$ to participate in the $t$-th round, with $|S^{(t)}| = C$.
We denote by $\tau$ the target number of training steps expected by the server.
However, as described previously, each client manages to perform only a portion of the requested computation, which we denote $\tau_i$ where $\cp_i \leq \cp$.
It corresponds to the computational budget of the $i$-th client, \ie, the number of local learning steps that this client is able to perform in the stipulated time window.

\textbf{\fedavg.} The default algorithm for federated training is \fedavg~\cite{mcmahan2017communication}, where the clients train on their local data using vanilla \sgd as the local optimizer.
Upon receiving the global model $\xsym^{(t, 0)}$ in the $t$-th communication round, the clients locally update it for $\tau_i$ steps as follows:
\begin{equation}
	\xsym_i^{(t, k+1)} \leftarrow \xsym_i^{(t, k)} - \clr . \gsym_i^{(t, k)}  \text{ where } k \in [0,\tau_i-1].  \label{eqn:fedavg_client}
\end{equation}
Here $\gsym_i^{(t, k)}$ is the gradient on a mini-batch of local data and $\clr$ is the learning rate, a hyperparameter received from the server.
At the end of the stipulated time window for round $t$, the client returns the final local model after $\tau_i$ steps, given by $\xsym_i^{(t, \tau_i)}$.
Recall that $\tau_i$ is typically different across clients, determined by the system characteristics of the respective client.
The server then aggregates the received models to produce the model for the next round as follows:
\begin{equation}
	\xsym^{(t+1, 0)} \leftarrow \sum_{i \in S^{(t)}} \frac{n_i}{n}  \xsym_i^{(t, \cp_i)},  \label{eqn:fedavg2} \\
\end{equation}
where the weights of the aggregation are defined by the amount of data held by the client $n_i$ and $n$ refers to the total data across all participating clients \ie $n = \sum_{i \in S^{(t)}} n_i$.

\subsection{Algorithms to address compute heterogeneity}
\label{subsec:algorithms_for_htr}

Several algorithms have been introduced to address systems heterogeneity in \ac{FL}.
These algorithms can be broadly classified into two groups: \emph{i)} optimization-based and \emph{ii)} model architecture based.
The optimization-based approaches modify the local training at the clients, the server's aggregation rule or both.
Prominent examples of this category include \fedprox~\cite{li2018federated} and \fednova~\cite{NEURIPS2020_564127c0}.
\fedprox~\cite{li2018federated} modifies the client's local training to include a penalty term that prevents the client's model from drifting away from the global model sent by the server.
This helps mitigate the discrepancy between the model updates of the faster clients and the slower clients.
\fednova~\cite{NEURIPS2020_564127c0} proposes a novel aggregation at the server which weighs model updates by accounting for the number of local steps conducted by the clients.

The model architecture based approaches such as \mbox{RAM-Fed}~\cite{10.1145/3580305.3599521}, \heteroFL~\cite{diao2021heterofl}, \adaptCL~\cite{zhou2021adaptcl} allow clients to train on heterogeneous or different model architectures.
This enables slow clients to train on smaller architectures and thus meet the reporting deadline of the server with the desired number of local steps.
The challenge here is to appropriately downsize the global model based on the client's system characteristics without compromising final accuracy.
\heteroFL proposes to shrink the \ac{DNN} model along the width dimension but retains the full depth for all clients.
\scalefl~\cite{ilhan2023scalefl} improves upon \heteroFL by shrinking along both the width as well as the depth dimensions of the \ac{DNN}, thereby maintaining high accuracy.
Additionally, \scalefl modifies the client local training to use self-distillation which prevents the performance drop caused by using a smaller \ac{DNN}.
On similar lines, the \aqfl algorithm~\cite{abdelmoniem:2021:heterogeneity} adapts quantization levels (bit-widths) of models on client devices instead of model architectures.
However, \aqfl faces several practical challenges like profiling of clients to accurately predict quantization level, thus limiting its adoption.

Our approach, \sys, is optimization-based and is orthogonal to the aforementioned approaches.
The guessing technique in \sys is generic enough to be applied in combination with \fedprox, \fednova but also \heteroFL, \scalefl, etc.
In \Cref{subsec:gel_against_sota_eval,subsec:gel_vs_scalefl}, we demonstrate the effectiveness of \sys in further boosting the performance of the above approaches.

\subsection{Hyperparameter tuning in \FL}
\label{subsec:hp_in_FL}
Selecting the right set of hyperparameters, such as batch-size, model architecture, learning rate, \etc is crucial to any \ac{ML} training pipeline.
This is typically achieved via random search, grid search, or adaptive search in centralized settings~\cite{bergstra12a:RS,hyperband}.
Due to the decentralized nature of data, selecting hyperparameters effectively in \ac{FL} is even more challenging~\cite{zhou2023singleshot,nips2021:fedex} where such search-based approaches prove extremely expensive in communication and computations costs.
In the \ac{FL} setting, FedEx~\cite{nips2021:fedex} has been proposed to improve over the search-based \ac{HP} tuning algorithms like Random Search~\cite{bergstra12a:RS} and Hyperband~\cite{hyperband} by leveraging the weight-sharing technique of Neural Architecture Search (NAS).
While FedEx can efficiently tune client local hyperparameters during the normal execution of \FL algorithms, it still requires search-based techniques mentioned above for some global hyperparameters like server-side learning rate.
The recently introduced \flora~\cite{zhou2023singleshot} algorithm achieves \ac{HPO} in one round by aggregating loss surfaces from clients.
Nevertheless, these approaches entail overheads which still might be prohibitive in practice.

\textbf{\fedopt framework.} Out of several hyperparameters, the learning rate is known to be particularly important~\cite{nar2018stepsize,charles2020outsized}.
Adaptive optimizers like \adam~\cite{kingma2017adam} and \yogi~\cite{NEURIPS2018_90365351} are less sensitive to learning rate tuning, thus representing robust alternatives to algorithms based on \ac{SGD}.
Reddi \etal~\cite{reddi2021adaptive} propose their federated equivalents under the general \fedopt framework which offers flexible choices of the client optimizer (\copt) and the server optimizer (\sopt).
More importantly, they study the sensitivity of several algorithms to the choices of client learning rate ($\clr$) and server learning rate ($\slr$) over a 2-D grid of values.
An algorithm is thus regarded as relatively \textit{easier to tune} if it produces good performance across several choices of parameter values.
We analyze \sys in the \fedopt framework and show that \sys improves performance with several $(\clr, \slr)$ combinations for the \fedavg and \fedyogi algorithms.
\sys can thus relax exhaustive tuning when the current methods remain expensive.

\subsection{Client momentum and its impact}
\label{subsec:utility_of_mom}
Although vanilla \sgd provides reasonable performance, the robust and fast convergence of momentum-based optimizers has been critical to the success of deep learning applications~\cite{pmlr-v28-sutskever13, NEURIPS2019_b8002139}.
Momentum is a technique used in optimization algorithms to accelerate convergence by smoothing out updates.
In simple words, it builds on the idea of incorporating a memory of past gradients to guide the current update direction.
As the gradients are computed, they are  accumulated in a velocity vector $(\vsym)$ as follows:
\begin{equation}
	\label{eqn:mom_on_clients}
	\vsym_i^{(t, k+1)} \leftarrow \alpha \vsym_i^{(t, k)} - \clr \gsym_i^{(t, k)} \text{ where } k \in [0,\tau_i-1].
\end{equation}
The initial value of the velocity is set to zero \ie $\vsym_i^{(t, 0)} = 0$ and the parameter $\alpha$ controls the weight given to the history in comparison to the current gradient.
At each step $k$, instead of using the current gradient $\gsym_i^{(t, k)}$ which can be noisy, the model is updated using the current velocity $\vsym_i^{(t, k+1)}$ which provides a more stable update as follows:
\begin{equation}
	\label{eqn:mom_model_update_on_clients}
	\xsym_i^{(t, k+1)} \leftarrow \xsym_i^{(t, k)} + \vsym_i^{(t, k+1)} \text{ where } k \in [0,\tau_i-1].
\end{equation}
The above optimizer is known as the \sgd with momentum optimizer (SGDM), contrasting with standard \sgd optimizer in \cref{eqn:fedavg_client}.

\begin{figure}[t]
	\includegraphics{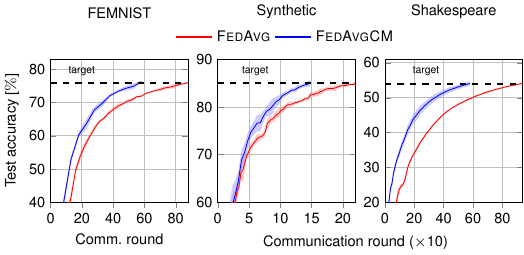}
	\caption{
		Comparison between \fedavg and \fedavgCM.
		The learning rate is tuned separately for both algorithms. Using client-side momentum achieves target accuracy nearly 2$\times$ faster.
	}
	\label{fig:vanilla_sgd}
\end{figure}

We refer to the version of the \fedavg algorithm that uses \sgdm as the \copt as \fedavgCM (federated averaging with client momentum).
To illustrate the performance difference when using momentum, we experiment and chart in Figure \ref{fig:vanilla_sgd} the learning curves for three different tasks using \fedavg and \mbox{\fedavgCM} algorithms.
Notably, using momentum speeds up convergence with respect to the vanilla version by nearly $2\times$ in communication rounds.
In this work, we exploit momentum to achieve client-side guessing in the face of limited computational budgets, as described next.

\section{\sys}
\label{sec:gel_algo}

In \ac{FL}, participating clients are often limited in their capacity to contribute to the training process.
We refer to this limitation by their computational budget $\cp_i$, which dictates the number of model update steps that the client performs when participating in a training round.
\textit{Given the computational budget constraints of clients $[\cp_1, \cp_2, \ldots, \cp_C]$ selected for a training round, the goal of \sys is to maximize the amount of progress made towards the optimal global model.}

\subsection{\sys: Guess and Learn Algorithm}
\label{subsec:gel_main}
\sys achieves its goal by \textit{guessing} future model update steps for every client.
The number of such updates is given by $[\cp'_1, \cp'_2, \ldots, \cp'_C]$.
Therefore, each client virtually performs \mbox{$[\cp_1+\cp'_1, \cp_2+\cp'_2,\ldots,\cp_C+\cp'_C]$} total learning steps, thereby boosting convergence.
These guessed updates do not require any extra gradient computation but only %
a model update step, a much faster computation than a forward plus a backward pass through the deep model.

In order to perform the guessing, \gel leverages momentum based optimizers on the client side.
In this work, we present our analysis and results using the \sgdm (\sgd with momentum) as the \copt, detailed previously in \Cref{subsec:utility_of_mom}.
While we focus on \sgdm for simplicity, we note that other choices such as \adam, \yogi, \etc are also possible.
On the server side, one is free to choose any \sopt.
Therefore, \sys can be flexibly combined with several federated algorithms, including \fedprox and \fednova.
\Cref{tab:list_of_algos} (\Cref{subsec:appendix_opt_list}) provides a complete list of all algorithms and their corresponding \sys versions along with the client and server optimizers used by each.
We detail next the procedure for guessed steps.

As discussed in \Cref{subsec:utility_of_mom}, upon training for $\tau_i$ local steps, clients produce $\xsym_i^{(t, \tau_i)}$ as the final local model.
At this point, the clients have exhausted their computational budgets $\cp_i$ and no more gradients can be computed.
Notably, the accumulated momentum $\vsym_i^{(t, \cp_i)}$ still contains useful information for the next update.
In the absence of the next gradient for step $\cp_i + 1$, the value of the gradient can be substituted with a proxy,
\begin{equation}
	\vsym_i^{(t, \cp_i+1)}  \leftarrow \alpha \vsym_i^{(t, \cp_i)} - \clr \gsym_i^{(proxy)}.
\end{equation}
In \sys, we use $\gsym_i^{(proxy)} = 0$.
This can be interpreted as deriving the subsequent update steps solely from the accumulated momentum.
We can repeat the same update for $k$ steps, resulting in
\begin{equation}
	\vsym_i^{(t, \cp_i+k)}  = \alpha \vsym_i^{(t, \cp_i+k-1)} \ldots = \alpha^k \vsym_i^{(t, \cp_i)}
	\label{eqn:recurrence_v}
\end{equation}
Assuming we repeat for $\cp'_i$ total steps, the model will be updated as follows:
\begin{align}
	\xsym_i^{(t, \cp_i+\cp'_i)}  &\leftarrow \xsym_i^{(t, \cp_i)}  + \sum_{k = 1}^{\cp'_i} \vsym_i^{(t, \cp_i+k)} \text{ \hspace{8mm} (from eqn.~\ref{eqn:mom_model_update_on_clients})}\nonumber\\
	 &\leftarrow \xsym_i^{(t, \cp_i)} + \left(\sum_{k=1}^{\cp'_i}\alpha^k\right) \vsym_i^{(t, \cp_i)} \text{\hspace{2mm} (from eqn.~\ref{eqn:recurrence_v})}\nonumber
\end{align}
The final model after $\cp'_i$ guessed steps is:
\begin{equation}
	\xsym_i^{(t, \cp_i+\cp'_i)} \leftarrow \underbrace{\xsym_i^{(t, \cp_i)}}_{\substack{\text{Last model before the} \\ \text{budget is exhausted}}}  + \underbrace{\left(\alpha \frac{1-\alpha^{\cp'_i}}{1-\alpha} \right) \vsym_i^{(t, \cp_i)}}_{\substack{\text{Nudge in the direction} \\ \text{ of momentum.}}}\\
	\label{eqn:gel_final}
\end{equation}
The guessed learning steps are effectively a nudge in the direction of momentum with the corresponding step size $\frac{\alpha(1-\alpha^{\cp'_i})}{1-\alpha}$ decided by the number of guessed steps $\cp'_i$.
More importantly, the number of \textit{guessed updates} $\cp'_i$ can be chosen differently for different clients, possibly depending on their actual steps performed $\cp_i$.
Recall from \Cref{sec:introduction} that the server requests a fixed number of learning steps $\cp$ in the stipulated time window.
One approach to establishing $\cp'_i$ is to assign it the value of the remaining work, \ie $\cp'_i = \cp - \cp_i$.
This strategy also homogenizes the total virtual work $\cp_i + \cp'_i$ across the clients.
We demonstrate the results with this strategy in \Cref{sec:exp_res}.

We provide the pseudocode of \sys in Algorithm~\ref{algo:GeL} and describe it below.
Similar to \fedavg,
the server selects a subset of clients and multicasts the global model $\xsym^{(t, 0)}$ along with the desired number of local steps $\cp$ for training in the current round (lines 4-7).
After initialization (lines 12-13), the clients train on their local data only up to their computational budget $\cp_i$ instead of $\cp$ (lines 14-18).
For the remaining undone steps $\cp-\cp_i$, the clients compensate by performing an equivalent number of guessed update steps $\cp'_i$ (lines 19-20).
Note that this is computed as a single operation (line 20) instead of iterative steps.
It also does not entail any gradient computations, hence is a cheap operation.
Finally, the server aggregates received model updates (line 8) and produces the new global model using a \sopt of its choice (line 9).
The \sopt could be the standard weighted aggregation of \fedavg or more advanced aggregation like \fednova as described in \Cref{sec:rw_and_background}.

\begin{algorithm}[t!]
	\caption{\sys. \label{algo:GeL}The $C$ selected clients are indexed by $i$; $\cp$ is the number of steps expected by the server; $\cp_i$ and $\cp'_i$ are the computational budget and number of guessed updates; $p_i$ is the weight; $\clr$ and $\slr$ refer to the client and the server learning rates, respectively. The server can choose any \sopt while the client must use an optimizer that accumulates the first moment of gradients (\sgdm shown below).}
	\DontPrintSemicolon
	\SetAlgoNoLine
	\small
	\SetKwFunction{ClientUpdate}{ClientUpdate}
	\textbf{Server Executes:}\; \Indp
	Initialise $\xsym^{(0,0)}$ \;
	\For{$t = 0,1,2,\dots$}
	{
		$S^{(t)} \leftarrow$ Server randomly selects $C$ clients\; %
		\For{each $i \in S^{(t)}$ in parallel}
		{
			$\Delta_i^{(t)} =$ ClientUpdate$\left( i, \xsym^{(t,0)}, \cp\right) $ \;
		}
		$\Delta^{(t)} = \sum_{i \in S^{(t)}} p_i \Delta_i^{(t)}$ \;
		$\xsym^{(t+1, 0)} = \text{\sopt}(\xsym^{(t, 0)}, -\Delta^{(t)}, \slr, t)$ \;
	}
	\Indm
	\BlankLine \BlankLine
	\textbf{ClientUpdate($i, \xsym^{(t,0)}, \cp$):} \Comment*[r]{Client $i$}
	\Indp
	$\vsym_i^{(t, 0)} = 0$ \Comment*[r]{initialize momentum}
	$\xsym_i^{(t,0)} = \xsym^{(t,0)}$ \Comment*[r]{initialize model}
	\Comment{Compute up to budget $\cp_i$}
	\For{step $k = 1$ \KwTo $\cp_i$}
	{
		Compute $\gsym_i^{(t, k-1)}$ \;
		$\vsym_i^{(t, k)} \leftarrow \alpha \vsym_i^{(t, k-1)} - \clr \gsym_i^{(t, k-1)}$ \;
		$\xsym_i^{(t, k)} \leftarrow \xsym_i^{(t, k-1)} + \vsym_i^{(t, k)}$ \;
	}
	\Comment*[l]{Guessed update step}
	Choose $\cp'_i = \cp-\cp_i$ \;
	$\xsym_i^{(t, \cp_i+\cp'_i)} \leftarrow \xsym_i^{(t, \cp_i)} + \alpha \frac{1-\alpha^{\cp'_i}}{1-\alpha}\vsym_i^{(t, \cp_i)}$ \;
	$\Delta_i^{(t)} \leftarrow \xsym_i^{(t, \cp_i+\cp'_i)} - \xsym_i^{(t, 0)}$ \;
	\BlankLine
	\KwRet{$\Delta_i^{(t)}$ to the server}

\end{algorithm}

\subsection{Convergence analysis of \sys.}
\label{subsec:convergence_analysis}
We now present the convergence result for the \mbox{\fedavgCMG} algorithm.
This result is based on the general theoretical framework of Wang \etal~\cite{NEURIPS2020_564127c0} which subsumes a suite of \FL algorithms whose accumulated local changes $(\xsym_i^{(t, \cp_i)} - \xsym_i^{(t, 0)})$ can be written as a linear combination of gradients.
More precisely, algorithms for which
$$\Delta_i^{(t)} =  \xsym_i^{(t, \cp_i)} - \xsym_i^{(t, 0)} = -\clr \Gsym_i^{(t)} \asym_i$$
where matrix $\Gsym_i^{(t)} = [\gsym_i^{(t, 0)}, \gsym_i^{(t, 1)},\ldots,\gsym_i^{(t, \tau_{i-1})}] \in  \mathbb{R}^{d \times \tau_i}$ stacks all local gradients and $\asym_i \in \mathbb{R}^{\tau_i}$ defines the coefficients of this linear combination are subsumed by the general theoretical framework.

Since the guessed step is derived solely out of the first moment of gradients (a linear combination), the accumulated updates in \mbox{\fedavgCMG} algorithm obey the above structure.
We derive the accumulated update and the gradient coefficient vector $\asym_i$ in an elaborate proof in \Cref{subsec:proof_of_lemma}, leading to:
\begin{equation*}
	\xsym_i^{(t, \cp_i+\cp'_i)} - \xsym_i^{(t, 0)} =-\clr\sum_{k=0}^{\tau_i-1} \left[\frac{1 -  \alpha^{\cp'_i + \cp_i - k}}{1-\alpha}\right] \gsym_i^{(t, k)}
\end{equation*}

\begin{lemma}[\textbf{Accumulated updates in \fedavgCMG}]
	\label{lem:linear_comb}
	When performing $\cp_i$ gradient steps and $\cp'_i$ guessed steps, the accumulated updates in the \fedavgCMG algorithm form a linear combination of gradients with the gradient coefficients, given by
	\begin{equation}
		\label{eqn:lemma_coeff_eqn}
		\asym_i = [1-\alpha^{\cp'_i+\cp_i}, 1-\alpha^{\cp'_i+\cp_i-1}, ..., 1-\alpha^{\cp'_i + 1}]/(1-\alpha)
	\end{equation}
\end{lemma}
Based on Lemma~\ref{lem:linear_comb} and the above theoretical framework, we show that \fedavgCMG algorithm converges at the rate $\mathcal{O}(1/\sqrt{m \bar{\cp} T})$, where $\bar{\cp} = \frac{1}{m} \sum_{i = 1}^m \cp_i$ and $T$ is the number of communication rounds (proof in \Cref{subsec:appendix_proof}).

\subsection{Discussion and insights}
\label{subsec:gel_insights}
The presented analysis serves two purposes: \emph{(i)} it confirms that the guessed updates in \sys do not jeopardize the convergence by showing that \fedavgCMG has a similar asymptotic convergence rate to standard \fedavg, with slightly different constants in the inequality; and \emph{(ii)} it provides interesting insights into understanding \sys.

\begin{figure}
	\includegraphics{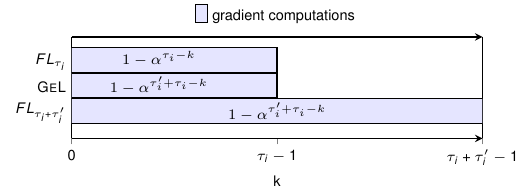}
	\caption{Gradient coefficients for \ac{FL} under $\cp_i$ and $\cp_i + \cp'_i$ local steps and \sys (only numerators shown, the denominator is $1-\alpha$ for all).
		$k$ denotes the local step number.
		While each client $i$ manages to conduct only $\cp_i$ steps due to system constraints, our analysis shows that \sys adapts the coefficients of their gradient combination as if it virtually conducted $\cp_i + \cp'_i$ steps, matching the coefficients of $FL_{\cp_i + \cp'_i}$.
	}
	\label{fig:GeL_coeffs}
\end{figure}

Guesses in \sys are derived out of gradients computed until the budget $\cp_i$ and accumulated in the momentum vector.
We showed that the final update in \sys forms a linear combination of gradients with the coefficients described in Eq.~\ref{eqn:lemma_coeff_eqn}.
We contrast these with the coefficients for two instances of the \fedavgCM algorithm (\ie, no \gel), one performing $\cp_i$ gradient-based steps, and the other  $\cp_i + \cp'_i$ gradient-based steps in \Cref{fig:GeL_coeffs}.

Interestingly, the coefficients in \gel correspond exactly to the coefficients that would have resulted for all gradients $\gsym_i^{(t,k)}$ with $k \leq \cp_i-1$, had the client performed all $\cp_i + \cp'_i$ steps as actual gradient-based steps.
In essence, \gel computes the sequential inter-dependent gradients in a regular way.
However, once computed, it cleverly combines them using coefficients that would have resulted if a greater number of actual gradients were computed.
This simple modification results in notable speedups, as  we demonstrate in \Cref{sec:exp_res}.

\subsection{Distributed execution of \sys}

\sys is explicitly designed to function within the standard distributed execution model of \ac{FL}, where a central server coordinates training across a population of decentralized and heterogeneous client devices. Below, we reiterate how \sys is integrated into this environment and executed in a distributed fashion.
At the beginning of each FL round, the server selects a subset of participating clients and transmits the current global model parameters. Each selected client then performs local training, consisting of: \emph{(i)} standard gradient-based updates on private data, and \emph{(ii)} additional lightweight guessed updates that require no gradient computation. These guessed updates are executed entirely locally and independently by each client, to compensate for the missed gradient-based updates.
Crucially, clients in \sys are stateless: clients reset momentum at the start of each round, eliminating the need for the server to manage or transmit momentum states.
This ensures no extra communication overhead beyond the standard \ac{FL} protocol.

After completing both gradient-based and guessed updates, clients send their updated model parameters back to the server. The server aggregates these updates using the chosen aggregation strategy (\eg, \fednova, \fedadam) to update the global model. This process repeats over multiple rounds until convergence.
By design, \sys maintains compatibility with widely-used optimizations and techniques in \ac{FL}, including: model compression~\cite{10272515,NEURIPS2018_33b3214d,8889996,10.5555/3524938.3525485}, differential privacy~\cite{9069945} and secure aggregation~\cite{45808,9336021}.
It operates entirely within the established constraints and assumptions of distributed \ac{FL}, introducing no additional coordination or communication overhead.
At the same time, \sys effectively addresses the limitations of resource-constrained clients and boosts \ac{FL} convergence, as we demonstrate next.

\section{Experimental results}
\label{sec:exp_res}

In this section, we aim to thoroughly evaluate the performance of \sys. We begin by detailing our experimental setup in \Cref{subsec:exp_setup}. We then address several key questions: How does \sys enhance the \fedavgCM baseline on the LEAF benchmark (\Cref{subsec:gel_vs_baseline_eval})? What performance gains can be achieved when combining \sys with different optimization-based approaches (\Cref{subsec:gel_against_sota_eval,subsec:eval_gel_in_fedopt})? How effectively does \sys improve the performance under untuned learning rate settings (\Cref{subsec:gel_untuned_eval,subsec:eval_gel_in_fedopt})? We also explore the guessing limit of \sys (\Cref{subsec:gel_infinite_guesses_eval}), the impact of budget range on its performance (\Cref{subsec:changing_budgets}), and its behavior on diverse tasks and larger models (\Cref{subsec:additional_datasets}). Finally, we evaluate the performance improvements when combing \sys with \scalefl, a model architecture-based approach (\Cref{subsec:gel_vs_scalefl})

\begin{table*}[th]
	\centering
	\caption{Summary of FL benchmarks used in this work.}
	\label{tab:benchmark_summary}
	\begin{tabular}{ c c c c r r r }
		\toprule
		\multirow{2}{*}{\textbf{Task}} & \multirow{2}{*}{\textbf{Dataset}} & \textbf{ML} & \multirow{2}{*}{\textbf{Model}} & \textbf{Total} & \textbf{Total} & \textbf{Target} \\
		& & \textbf{Technique} & & \textbf{Clients} & \textbf{Samples} & \textbf{Accuracy} \\
		\toprule
		Image  & \multirow{2}{*}{FEMNIST} & \multirow{2}{*}{CNN} & 2 Conv2D & \multirow{2}{*}{\num{3597}} & \multirow{2}{*}{\num{734463}} & \multirow{2}{*}{\SI{76}{\%}} \\
		Classification &  &  & Layers & & & \\
		\midrule
		Cluster & \multirow{2}{*}{Synthetic} & Traditional & Logistic & \multirow{2}{*}{1000} & \multirow{2}{*}{\num{96374}} & \multirow{2}{*}{\SI{85}{\%}} \\
		Identification & & ML & Regression & & & \\
		\midrule
		Next Word &  \multirow{2}{*}{Shakespeare} &  \multirow{2}{*}{RNN} & Stacked & \multirow{2}{*}{\num{660}} & \multirow{2}{*}{\num{3678451}} & \multirow{2}{*}{\SI{54}{\%}} \\
		Prediction & & & LSTM & & & \\
        \midrule
        Image &  \multirow{2}{*}{CIFAR-10/100} &  \multirow{2}{*}{CNN} & ResNet/ & \multirow{2}{*}{$20-200$} & \multirow{2}{*}{\num{50000}} & \multirow{2}{*}{\SI{80}{\%}} \\
        Classification & & & MSDNet & & & \\
        \midrule
        Text &  \multirow{2}{*}{\agnews} &  \multirow{2}{*}{Transformer} & \multirow{2}{*}{DistilBert} & \multirow{2}{*}{\num{100}} & \multirow{2}{*}{\num{120000}} & \multirow{2}{*}{\SI{85}{\%}} \\
        Classification & & & & & & \\
		\bottomrule
	\end{tabular}
\end{table*}

\subsection{Experimental setup}
\label{subsec:exp_setup}

\textbf{Datasets and data partitioning.}
We evaluate \sys and the baselines on several different learning tasks -- image classification on the FEMNIST, \cifar and \cifarH datasets, text generation on the Shakespeare dataset, cluster identification on the Synthetic dataset and lastly, text classification on the \agnews dataset.
The FEMNIST, Shakespeare, and Synthetic datasets are taken from the LEAF benchmark~\cite{caldas2018leaf} for \FL, used in several previous works~\cite{mcmahan2017communication,li2018federated,reddi2021adaptive,charles2021large}.
These datasets also exhibit a natural non-IID partitioning, \eg each writer is a separate client in the FEMNIST dataset.
Similarly, in the Shakespeare dataset, each speaking role from the plays of William Shakespeare represents a distinct client, capturing the heterogeneity in vocabulary and speaking style inherent to different characters.
Consequently, the total number of clients is also predetermined in the LEAF benchmark, for instance, the total number of speaking roles in the Shakespeare dataset.
For the CIFAR-10/100~\cite{krizhevsky2009learning} and \agnews~\cite{zhang2015character} datasets, we use Dirichlet distribution ($Dir(.)$) with different concentration parameters $\{0.1, 1, 100\}$ to control the amount of data heterogeneity, as done in several previous works~\cite{ilhan2023scalefl,NEURIPS2020_564127c0}.
For these datasets, we conduct experiments varying the number of clients from \num{20} to \num{200}.

\textbf{Models.}
For the tasks from the LEAF benchmark, we use the same models as McMahan \etal~\cite{mcmahan2017communication} and Li \etal~\cite{li2018federated}.
For \cifar and \cifarH, we experiment with different models, including ResNet-8~\cite{he2016deep}, ResNet-110~\cite{he2016deep} and MSDNet~\cite{huang2018multiscale}.
For the \agnews dataset, we fine-tune DistilBert~\cite{sanh2019distilbert}, a transformer-based model, as done in previous work~\cite{sattler2021cfd}.
Note that in our experimental setup, we specifically focus on constrained edge settings, where resources such as memory and processing power are limited.
In this context, the selected models are well-suited and align with these limitations.
Table~\ref{tab:benchmark_summary} summarizes the learning tasks, datasets, and models.

\textbf{Hyperparameters.}
We tune the client learning rate $(\clr)$ for the \fedavgCM baseline for every dataset.
\fedprox, \fednova, \scalefl, and \sys use the same tuned learning rate for fairness.
The \fedavg algorithm in \Cref{fig:vanilla_sgd} has a separately tuned client learning rate due to the absence of client momentum.
Default server learning rate $\slr = 1$ is used in all experiments except for the \fedopt framework (\Cref{subsec:eval_gel_in_fedopt}), where we tune both the client and server learning rates $(\clr, \slr)$.
The number of selected clients $(C)$ per round is fixed at $20$ in all experiments, except the ones involving the \agnews dataset and \scalefl, where we use $C=10$.
Batch sizes of $5$, $20$, $20$, $16$ and $8$ are used for the Synthetic, FEMNIST, Shakespeare, \cifar/100, and \agnews datasets, respectively.
We set the momentum parameter ($\alpha$) to 0.9 and use $\beta_1 = 0.9$ and $\beta_2 = 0.99$ for the \yogi optimizer.
The adaptivity parameter of the \yogi optimizer is fixed at $10^{-3}$.
More details on hyperparameter tuning are in \Cref{subsec:appendix_hparams}.

\textbf{Client budgets}
Following Li \etal~\cite{li2018federated}, we perform the budget assignment every round in two steps: (1) participating clients are selected uniformly at random; and (2) each selected client uniformly randomly samples budgets from a range $\cp_i \in [a,b]$.
This resembles realistic \FL settings where client budgets are not only heterogeneous but may also vary across rounds for any individual client.
We set the server's expected $\cp$ to $b+5$.
This means that the server expects each client to perform $b+5$ local update steps.
The budget ranges ($\tau_i$) along with the desired value of $\tau$ are stated on top of the charts (\Cref{fig:gel_vs_fedavg}) per dataset.

\textbf{Metrics.}
We evaluate the performance of \sys along four metrics: top-1 test accuracy, speedup, network savings, and the number of gradient computations.
The performance speedup is measured in rounds of communication to achieve a predefined target accuracy.
We set these accuracy targets similar to the ones used in previous works ~\cite{mcmahan2017communication,caldas2018leaf,abdelmoniem:2021:heterogeneity}.
The network savings, presented alongside speedups, highlight the reduced communication costs.
Finally, we measure the evolution of test accuracy against the cumulative number of gradients computed by all participating clients until the target accuracy has been reached.
This metric specifically showcases the role of \sys in maximizing progress under stringent computational budget constraints.
We run each experiment with 5 random seeds and present the average values and the 95\% confidence interval.

\textbf{Implementation.}
All experiments were conducted on a compute cluster comprising 2 x AMD EPYC 7543 32-Core 2.8GHz CPU Processor, equipped with 8 x NVIDIA A100 SXM4 80GB GPU.
Our code, written in Python v3.9, is built on top of the TensorFlow~\cite{10.5555/3026877.3026899} and PyTorch~\cite{10.5555/3454287.3455008} frameworks and runs in a fully distributed fashion, with each participating client simulated as a separate operating system (OS) process.
We use \texttt{torch.distributed} for distributed communication and synchronization.
We will provide an open-source implementation of \sys for reusability and reproducibility.

\subsection{\sys against \fedavgCM on the LEAF benchmark}
\label{subsec:gel_vs_baseline_eval}

\begin{figure*}[t!]
	\centering
	\includegraphics{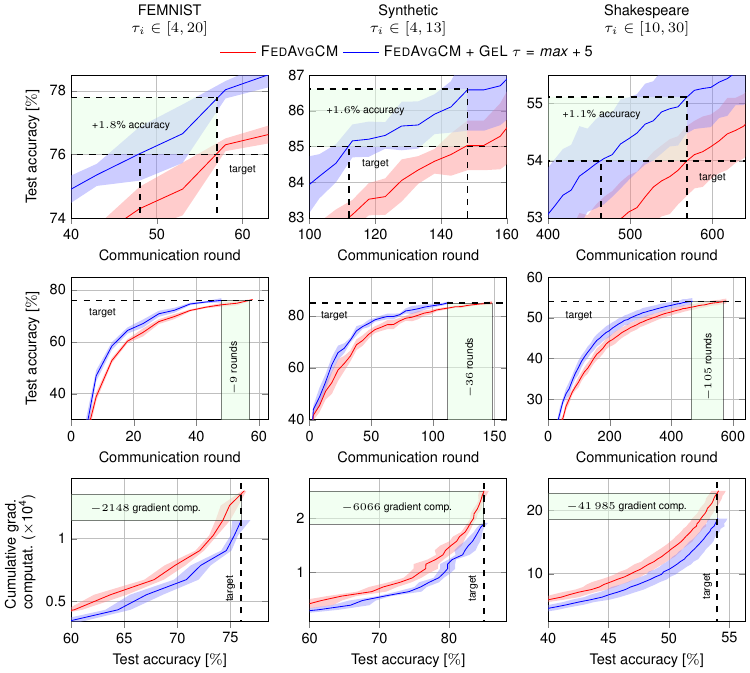}
	\caption{
Performance of \sys against \fedavgCM across datasets. \sys reaches the target accuracy faster than \fedavgCM on all the datasets (row-2).
\sys is also able to reach a better accuracy when \fedavgCM reaches the target accuracy (row-1).
Needing fewer communication rounds to reach the target accuracy, \sys performs significantly fewer gradient computations when compared to \fedavgCM (row-3).
	}
	\label{fig:gel_vs_fedavg}
\end{figure*}
\begin{table*}[ht]
	\centering
		\caption{\label{tab:gel_vs_fedavg}Performance comparison of \sys to \fedavgCM. \sys reaches the target accuracy up to 20-38\% faster in communication rounds in both tuned and untuned scenarios. In addition, we report the economy of resources in \sys in terms of network savings.}
	\begin{tabular}{ccSlccccc}
		\toprule
		\multirow{2}{*}{Dataset} & Target  &  {\multirow{2}{*}{$\clr$}} & & \multicolumn{2}{@{} c}{Comm. rounds until target accuracy}  &  \multirow{2}{*}{Speed up} & \# Model  & Network \\
		& accuracy & & & \fedavgCM & \sys & & parameters & savings \\
		\midrule
		\multirow{2}{*}{FEMNIST} & \multirow{2}{*}{76\%} & 0.02 & tuned & 57 & 48 & 18.8\% & \multirow{2}{*}{\num{6.1e5}} &  \SI{1.08}{\gibi\byte}\\
		& & 0.01 & untuned &95 & 69 & 37.7\% & & \SI{3.11}{\gibi\byte}\\
		\midrule
		\multirow{2}{*}{Synthetic} & \multirow{2}{*}{85\%} & 0.01 & tuned & 148 & 112 & 32.1\% & \multirow{2}{*}{\num{305}} & \SI{0.58}{\mebi\byte}\\
		& & 0.005 & untuned & 176 & 135 & 30.4\% & & \SI{0.66}{\mebi\byte}\\
		\midrule
		\multirow{2}{*}{Shakespeare} & \multirow{2}{*}{54\%} & 0.3 & tuned & 569 & 464 & 22.6\% & \multirow{2}{*}{\num{8.2e5}} & \SI{3.73 }{\gibi\byte}\\
		& & 0.15 & untuned & 820 & 639 & 28.3\%& & \SI{6.44}{\gibi\byte}\\
		\bottomrule
	\end{tabular}
\end{table*}

As noted in \Cref{subsec:utility_of_mom}, momentum provided a substantial speedup across all learning tasks.
Consequently, we consider the \fedavgCM algorithm as the baseline instead of the standard \fedavg algorithm throughout our experiments. We refer to the application of \sys on \fedavgCM as \fedavgCMG (\Cref{tab:list_of_algos}, \Cref{subsec:appendix_opt_list}).
Figure \ref{fig:gel_vs_fedavg} presents the performance results, with rows 1 and 2 showing the test accuracy versus communication rounds, and row 3 illustrating the evolution of test accuracy with respect to the total gradients computed by the system.

As described in \Cref{sec:gel_algo}, \sys performs  $\cp'_i = \cp - \cp_i$ guessed updates for every client.
By just performing these free learning steps, \sys speeds up convergence to target accuracy by up to $30\%$ in rounds of communication (speedup column of \Cref{tab:gel_vs_fedavg}).
Notably, even for the challenging Shakespeare dataset, \sys requires over $100$ fewer rounds to converge.
Furthermore, when both \sys and \fedavgCM reach the target accuracy, \sys achieves higher accuracy in all learning tasks, with a notable increase of approximately $2\%$ for the FEMNIST dataset (row-1 of \Cref{fig:gel_vs_fedavg}).
In terms of computation, \sys consistently requires significantly fewer gradient computations compared to \fedavgCM to achieve equal accuracy across all learning tasks. This translates to thousands of saved computations, reaching up to \num{42000} for the Shakespeare dataset (row-3 of \Cref{fig:gel_vs_fedavg}).

We also demonstrate that \sys is not very sensitive to large values of $\cpp_i$ by setting $\cpp_i = \infty$ in \Cref{subsec:gel_infinite_guesses_eval}.
Additionally, we assess \sys under different budget ranges in \Cref{subsec:changing_budgets}.

\subsection{\sys applied to \fedprox and \fednova}
\label{subsec:gel_against_sota_eval}

The \fedprox and \fednova algorithms were designed to address compute heterogeneity (\Cref{subsec:algorithms_for_htr}).
We investigate their combination with \sys and examine whether \sys can accelerate their convergence.

\begin{table*}[t]
	\centering
	\caption{
		\sys enhances existing \FL algorithms by achieving earlier target accuracy (\textit{Speedup}) and surpassing their accuracy (\textit{\sys accuracy beyond target}) when measured at the round when the baseline (\ie \fedprox or \fednova) achieves the target.
	}
	\begin{tabular}{c | c c c c | c c c c}
		\toprule
		\multirow{5}{*}{Datasets}	& \multicolumn{4}{c}{\fedprox}  & \multicolumn{4}{c}{\fednova} \\
		\cmidrule(lr){2-5}\cmidrule(lr){6-9}
		& \multicolumn{2}{c}{Comm. rounds until} & \multirow{4}{*}{Speedup} & GeL & \multicolumn{2}{c}{Comm. rounds until} & \multirow{4}{*}{Speedup} & GeL\\
		& \multicolumn{2}{c}{target accuracy} &  &  accuracy &  \multicolumn{2}{c}{target accuracy} &  & accuracy \\
		\cmidrule(lr){2-3}\cmidrule(lr){6-7}
		& \multirow{2}{*}{Default} & Default &  & beyond & \multirow{2}{*}{Default} & Default &  & beyond  \\
		& & + \sys &  & target [\%] & & + \sys &  & target [\%] \\
		\midrule
		FEMNIST $\cp_i \in [4,20]$ & 49 & 38 & 28.9\% & +2.19 & 63 & 55 & 11.5\% & +1.04 \\
		Synthetic $\cp_i \in [4,13]$ &157 & 112 & 40.2\% & +1.18 & 118 & 103 & 14.6\% & +0.94  \\
		Shakespeare $\cp_i \in [10,30]$ & 605 & 478 & 26.6\% & +1.26 & 578 & 478 & 20.9\% & +1.17  \\
		\bottomrule
	\end{tabular}
	\label{tab:cifar}
\end{table*}

\textbf{\fedprox + \sys.}
The gradients in the \fedprox algorithm account for the proximal penalty along with the regular loss function.
This being the only difference to \mbox{\fedavgCM}, guessed updates in \sys (Eq.~\ref{eqn:gel_final}) can be directly applied on top of \fedprox algorithm.
Table~\ref{tab:cifar} summarizes the results under the column \fedprox.
\fedprox + \sys takes significantly fewer communication rounds to reach the same target accuracy as \fedprox, achieving between $26$ and $40\%$ speedup across the learning tasks.
In addition, \fedprox + \sys reaches up to $2.19\%$ higher test accuracy by the round when default \fedprox achieves target accuracy.
These findings show that \sys can be seamlessly combined with \fedprox to boost the convergence of the latter.

\textbf{\fednova + \sys.}
\fednova normalizes the client model updates by considering the number of steps conducted by local clients (\Cref{subsec:algorithms_for_htr}).
Similar to \fedprox, \sys can also be easily applied on top of the \fednova algorithm.
We compare against the version of \fednova which uses the \sgdm optimizer, noting its superior performance~\cite{NEURIPS2020_564127c0}.
Table~\ref{tab:cifar} summarizes the results under column \fednova.
With a simple modification resulting from guessed updates, \sys speeds up default \fednova  by 10-20\%.
Moreover, it reaches nearly 1\% higher target accuracy for all datasets by the round when default \fednova reaches the target accuracy.
Therefore, \sys can be effectively combined with \fednova to accelerate convergence.

\subsection{\sys in untuned learning rate settings}
\label{subsec:gel_untuned_eval}

\begin{figure}[t]
	\centering
	\includegraphics{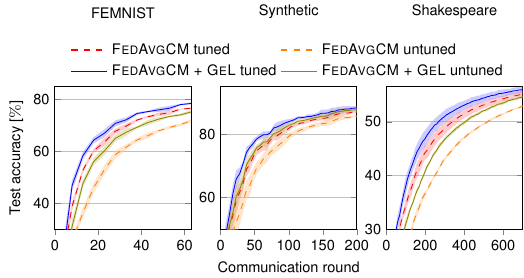}
	\caption{
 Impact of \sys on performance with tuned and untuned learning rate ($\clr$).
 \sys boosts convergence even when using a suboptimal learning rate.}
	\label{fig:tuned_vs_untuned}
\end{figure}

As motivated in \Cref{sec:introduction,sec:rw_and_background}, tuning learning rates is an arduous task in \FL.
Interestingly, \sys presents an alternative to exact tuning: gradients can be computed using \textit{untuned} learning rates, but combined using coefficients that maximize progress (\Cref{fig:GeL_coeffs}).
To demonstrate this, we consider \sys with two sets of learning rate values: the best and a non-best learning rate that still achieves the target accuracy.
We set the non-best learning rate to a value equal to half of the best one, although in practice this could be arbitrary.

\Cref{fig:tuned_vs_untuned} and \Cref{tab:gel_vs_fedavg} demonstrate the results.
In the untuned setting, \sys converges nearly 38\% faster for the FEMNIST dataset.
For the Synthetic dataset, \sys converges in even fewer rounds than the tuned \fedavgCM baseline.
Finally, for the challenging Shakespeare dataset, the speedup rises from 22\% in the tuned case to 28\% in the untuned case, taking over 150 fewer communication rounds to reach the same accuracy.
Intuitively, \sys exhibits this behavior because lower learning rates provide more room for improvement through guessing.
In other words, when step sizes are small, learning can smoothly progress in the direction of momentum, which \sys precisely exploits.
Furthermore, this speedup leads to nearly double the network savings in data volume compared to the tuned case (for FEMNIST and Shakespeare datasets), as indicated in Table~\ref{tab:gel_vs_fedavg}.
These results demonstrate that \sys achieves significant performance boosts without requiring perfectly tuned learning rates, making it a cost-effective alternative to expensive tuning.

\subsection{\sys in \fedopt framework}
\label{subsec:eval_gel_in_fedopt}

\begin{figure}[t]
	\centering
	\includegraphics{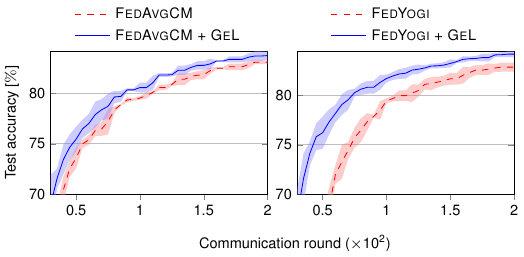}
	\caption{
\sys in the \fedopt framework.
Both the \fedavgCM and \fedyogi algorithms achieve higher accuracy in a given number of rounds when combined with \sys.
 }
	\label{fig:gel-vs-yogifedavg}
\end{figure}

In this section, we analyze the impact of server-side optimization on \sys performance.
We evaluate \fedavgCM and \fedyogi algorithms on the FEMNIST dataset with tuned client ($\clr$) and server learning ($\slr$) rates, following the procedure outlined in~\cite{reddi2021adaptive}.
See \Cref{subsec:appendix_hparams} for more details on tuning.
We observe in \Cref{fig:gel-vs-yogifedavg} that
the guessing mechanism in \sys continues to expedite empirical convergence, leading to higher accuracy within a fixed number of communication rounds.
Furthermore, algorithms in the \fedopt framework achieve better accuracies than previously (\Cref{fig:gel_vs_fedavg}), emphasizing the benefits of tuning both client and server learning rates.
This however comes at the significant cost of tuning.

\begin{figure}[t]
	\includegraphics{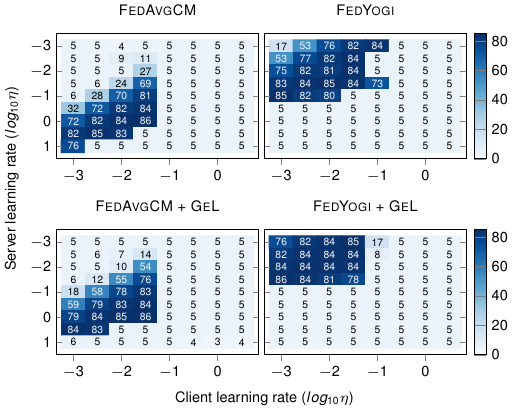}
	\caption{
            Final test accuracy achieved by \fedavgCM, \fedyogi, and their \sys counterparts on the grid of $(\clr, \slr)$ values for the FEMNIST dataset.
            \sys improves the test performance for a large set of parameter values, easing the effort of exhaustive tuning.
            While it hurts the performance of a few combinations, these parameter values are arguably not a safe choice being on the extreme of the spectrum.
        }
	\label{fig:gelimproveshparam}
\end{figure}

As previously established, we deem an algorithm as \textit{easy to tune} in case it produces good performance across several choices of parameter values.
\sys, in particular, through guessing updates can restore the performance of bad parameter values.
To justify this, we chart in Figure~\ref{fig:gelimproveshparam} the test accuracy grids upon running $1000$ rounds of training on the FEMNIST dataset.
Note that \sys improves the test performance for many $(\clr, \slr)$ values, providing good performance over a large set.
This is especially evident for the \fedyogiG combination.
While it hurts the performance of a few combinations of $(\clr, \slr)$, we argue that these parameters are the ones with very high learning rate values.
Hence, they are not the safest choice as they are susceptible to overshooting and divergence.
In conclusion, \sys alleviates the strong need for exhaustive tuning, serving as a practical alternative.

\subsection{What happens when guessing to the limit?}
\label{subsec:gel_infinite_guesses_eval}

\begin{figure}[t]
	\centering
	\includegraphics{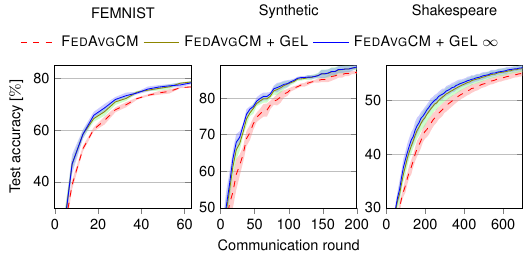}
	\caption{\sys with infinite guessed updates.
		We observe that guessing too much does not hurt convergence.
		This is because setting the number of guessed updates to $\infty$ yields a constant nudge parameter.
		Consequently, \sys does not require tuning the number of guesses.}
	\label{fig:toomuch}
\end{figure}

The initial motivation of \sys is to compensate for resource-constrained clients that are not able to compute as many learning steps as requested by the server.
Thus, in our experiments, we always set the number of guessed updates $\cp'_i = \cp - \cp_i$.
Doing so also equalized the amount of (virtual) total work across nodes, with part of this work done by gradientless steps.
One might wonder what would be the consequence if the clients guessed too many steps.
To answer this, by rewriting the nudge in \cref{eqn:gel_final},
$$
\frac{\alpha }{1-\alpha} - \frac{\alpha^{\cp'_i+1}}{1-\alpha}
$$
we observe that the guessed updates $(\cp'_i)$  only impact an exponentially decreasing term with parameter $\alpha < 1$.
Hence, performing a large number of guessed updates will not produce a significantly different nudge parameter compared to a small finite number.
In fact, setting $\cp'_i = \infty$ turns the second term to zero, yielding a constant step size of $\alpha/(1-\alpha)$.
Figure~\ref{fig:toomuch} presents the learning curves for \sys with an infinite number of guesses, alongside the \sys and \fedavgCM baselines from Figure~\ref{fig:gel_vs_fedavg}.
They confirm that \sys performs similarly with such a large number of guessed updates, corroborating our theoretical justification.
In essence, one can set a number of guessed updates equal to the compensatory number $\cp'_i = \cp - \cp_i$ or let all clients guess infinite steps.
This finding reveals that \sys does not require exhaustive tuning for the number of guessed updates, as both the above values tend to work well in practice.

\subsection{How does the budget range impact \sys?}
\label{subsec:changing_budgets}

We now address the incidental question of how client budget ranges affect the performance of \sys.
Intuitively, in order for guessing to be effective, \sys needs the clients to have accumulated local momentum through at least some local steps.
Hence, in an extreme case where clients do only one local step, \sys would be no better than the baseline.
Similarly, at the other extreme, where all clients successfully complete the expected amount of work, \sys offers limited improvements due to the lack of room for further optimization.
However, in the more realistic average case, we show that \sys is effective in boosting the baseline.

\begin{figure}[t]
	\centering
	\includegraphics{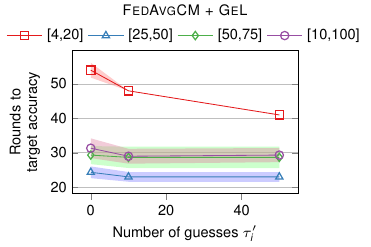}
	\caption{
		Impact of varying client budget ranges on \sys's performance on the FEMNIST dataset.
		$\cp'_i = 0$ yields the baseline \fedavgCM algorithm.
		\sys speeds up convergence in the scenarios with heterogeneous budgets, most notably in the stringent ($\tau_i \in [4, 20]$) case compared to the high budget ($\tau_i \in [50, 75]$)  and wide-ranging ($\tau_i \in [10, 100]$) scenarios.
		}

	\label{fig:changing-budget}
\end{figure}

We empirically confirm this by running the experiment with different ranges including $\tau_i \in [4, 20], [25, 50], [50, 75],$ and a wider range $[10, 100]$ on the FEMNIST dataset.
We vary the number of guesses $\cp'_i \in \{0, 10, 50\}$ and chart the rounds to target accuracy in \Cref{fig:changing-budget}.
When $\cp'_i=0$, we get the baseline.
As we increase $\cp'_i$, we control the effect of \sys.
Encouragingly enough, even a large number of guesses does not lead to divergence.
We observe that \emph{(i)} under stringent budget conditions \ie $[4, 20]$, \sys speeds up convergence from $54$ (baseline with $0$ guesses) to $42$ rounds;
\emph{(ii)} when the budgets increase to $[25, 50]$, \sys still brings speed up but its relatively lower than $[4, 20]$;
\emph{(iii)} when the budgets are too high $[50, 75]$, both the baseline and \sys suffer from client drift needing more rounds than the $[25, 50]$ case.
\sys, however, does not worsen the baseline.
Finally, we note that stringent resource constraints are likely to induce low-budget clients (\eg $\cp_i \in [4, 20]$) where \sys brings the most speed up.

\subsection{Performance on additional datasets}
\label{subsec:additional_datasets}
We extend our evaluations to more complex datasets and models, comparing \sys against \fedavgCM on the \cifar, \cifarH and the \agnews datasets.

\begin{table}[b]
	\centering
	\caption{\agnews dataset.
		\sys provides $13.8-15.8\%$ speedup in communication rounds to achieve $85\%$ target accuracy.
		 }
	\label{tab:ag_news}
	\resizebox{0.48\textwidth}{!}{
		\begin{tabular}{c  c c  c c}
			\toprule
			\multirow{2}{*}{Data htr.} & \multicolumn{2}{c}{Rounds to target acc. (\num{85}\%)} &  \multirow{2}{*}{Speedup} & \sys accuracy \\
			\cmidrule{2-3}
			& \fedavgCM & \fedavgCMG & & beyond target [\%] \\
			\midrule
			$Dir(1)$ & $330$ & $290$ & $13.8\%$ & $+0.87$ \\
			$Dir(100)$ & $220$ & $190$ & $15.8\%$ & $+1.14$\\
			\bottomrule
	\end{tabular}}
\end{table}

\subsubsection{\cifar and \cifarH datasets}
\label{subsubsec:cifar_eval}
\begin{figure}[t]
    \centering
    \includegraphics{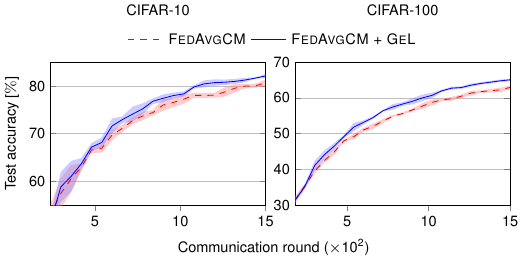}
    \caption{Performance on the \cifar and \cifarH datasets under $Dir(0.1)$.
    	\sys achieves $1.5\%$ and $2\%$ higher accuracy points on the \cifar and \cifarH datasets respectively in $1500$ communication rounds.
    }
    \label{fig:cifar10_cifar100}
\end{figure}

We consider a setup with $m = 50$ total clients with heterogeneously distributed data given by $Dir(0.1)$.
We use the ResNet-8 model and sample the client computational budgets as $\cp_i \in [5, 15]$.
We set $\cp'_i = 20$ and the total communication rounds to $1500$.
\Cref{fig:cifar10_cifar100} depicts the results where we observe that guessed updates in \sys effectively boost the convergence for both the \cifar and the \cifarH datasets.
After $1500$ communication rounds, \fedavgCMG achieves about $1.5\%$ higher absolute accuracy for the \cifar dataset and about $2\%$ higher absolute accuracy for the \cifarH dataset.

\subsubsection{\agnews dataset}
We consider a setup with $m=100$ clients and the task of fine-tuning the DistilBert model on the \agnews dataset, the largest model in our experimental bench comprising $\approx 67$ million parameters.
The client computational budgets are sampled from $\cp_i \in [10, 25]$ and we set $\cp'_i = 5$.
Results are shown in \Cref{tab:ag_news} for two different heterogeneity levels.
We observe that \sys enables $13.8-15.8\%$ speedup in communication rounds to reach the target accuracy of $85\%$.
This amounts to $30-40$ rounds of transfers being saved, which are significant for a model of the size of DistilBert.
Additionally, \sys achieves up to $1.14\%$ higher accuracy points at the round when the baseline reaches the target, further demonstrating the effectiveness of \sys.

\begin{table}[b]
    \centering
    \caption{Scalability study on the \cifar dataset. We run up to $1500$ rounds and report the round in which the target accuracy was achieved. \fedavgCM does not reach the target in $1500$ rounds when $\#$clients$=20$.}
    \label{tab:cifar10_scalability}
    \resizebox{0.48\textwidth}{!}{
    \begin{tabular}{c  c c  c c}
        \toprule
         \multirow{2}{*}{\#Clients} & \multicolumn{2}{c}{Rounds to target acc. ($80\%$)} &  \multirow{2}{*}{Speedup} & \sys accuracy \\
         \cmidrule{2-3}
         & \fedavgCM & \fedavgCMG & & beyond target [\%] \\
         \midrule
         $20$ & $1500+$ & $1320$ & $13.6\%$ & $+1.03$ \\
         $50$ & $1380$ & $1140$ & $21.1\%$ & $+1.27$\\
         $100$ & $1320$ & $1140$ & $15.8\%$ & $+0.87$\\
         $200$ & $1380$ & $1200$ & $15.0\%$ &  $+1.65$ \\
         \bottomrule
    \end{tabular}}
\end{table}

\subsubsection{Scalability study on the \cifar dataset}
In this section, we assess the scalability of \sys by varying the total number of clients in the federated system from $20$ to $200$.
We consider the \cifar dataset and keep the remaining setup the same as \Cref{subsubsec:cifar_eval}.
\Cref{tab:cifar10_scalability} reports the number of communication rounds required to achieve a target accuracy of $80\%$.
\sys reaches the target accuracy $180-240$ rounds earlier across various network sizes, consistently delivering a speedup of $13.6 – 21.1\%$.
For the specific case of $m=20$ clients, where the baseline fails to reach the target accuracy even after $1500$ rounds, \sys successfully achieves it in just $1320$ rounds.
Thus, \sys provides consistent boosts even under growing network sizes.

\begin{figure*}[t]
    \centering
    \includegraphics{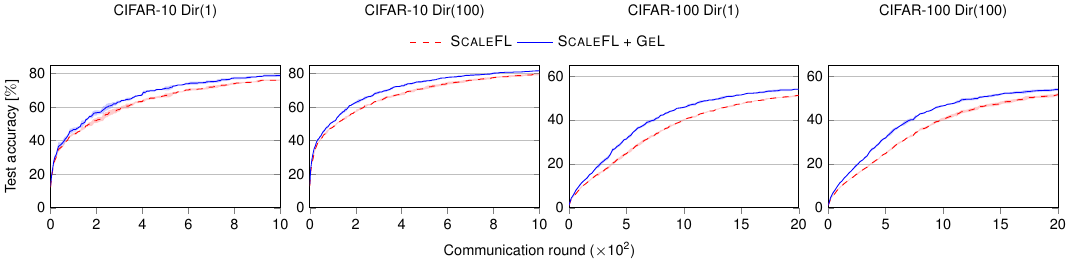}
    \caption{\scalefl vs \sys on the \cifar and \cifarH datasets using the MSDNet model architecture.
    		\sys speeds up \scalefl, achieving nearly $2\%$ higher accuracy points in each scenario.
    		The corresponding final accuracy values are reported in \Cref{tab:gel_vs_scalefl}.
    }
    \label{fig:gel_vs_scalefl}
\end{figure*}

\subsection{\sys combined with \scalefl}
\label{subsec:gel_vs_scalefl}
So far, we have demonstrated the effectiveness of \sys in combination with the optimization-based approaches for compute heterogeneity such as \fedprox, \fednova, \fedyogi, \etc.
The second set of approaches involve adjusting the model architecture size assigned to different clients depending on their resource constraints.
Prominent examples include \heteroFL, \adaptCL, \scalefl, \etc as discussed in \Cref{subsec:algorithms_for_htr}.
While these approaches address resource heterogeneity, they are orthogonal to \sys and can be effectively combined with \sys to further boost convergence.
To illustrate this, we combine \sys with \scalefl, the current state-of-the-art and present the results in \Cref{tab:gel_vs_scalefl} and \Cref{fig:gel_vs_scalefl}.

\begin{table*}[t]
    \centering
    \caption{\scalefl combined with \sys. We report the accuracy obtained after \num{1000} and \num{2000} communication rounds for \cifar and \cifarH datasets respectively. \sys boosts \scalefl, achieving $2-5\%$ higher accuracy points across different combinations of model architectures and data heterogeneity values.}
    \label{tab:gel_vs_scalefl}
    \begin{tabular}{c c c c c c}
        \toprule
        \multirow{2}{*}{Model} & \multirow{2}{*}{Algorithm} & \multicolumn{2}{c}{\cifar} & \multicolumn{2}{c}{\cifarH} \\
         \cmidrule(lr){3-4}\cmidrule(lr){5-6}
         & & $Dir(1)$ & $Dir(100)$ & $Dir(1)$ & $Dir(100)$ \\
         \midrule
         \multirow{2}{*}{ResNet110} & \scalefl & $72.01 {\scriptstyle \pm 0.49}$ & $77.39 {\scriptstyle \pm 0.20}$ & $38.71 {\scriptstyle \pm 0.51}$ & $39.32 {\scriptstyle \pm 1.06}$ \\
         & \scaleflg & $\bm{74.90} {\scriptstyle \pm \bm{0.24}}$
         & $\bm{79.40} {\scriptstyle \pm \bm{0.83}}$ & $\bm{43.58} {\scriptstyle \pm \bm{0.46}}$ & $\bm{45.15} {\scriptstyle \pm \bm{0.47}}$\\
         \midrule
            \multirow{2}{*}{MSDNet24} & \scalefl & $76.16 {\scriptstyle \pm 0.30}$ & $79.57 {\scriptstyle \pm 0.65}$ &
            $51.24 {\scriptstyle \pm 0.29}$ &
            $51.53 {\scriptstyle \pm 0.76}$\\
            & \scaleflg & $\bm{78.75} {\scriptstyle \pm \bm{0.76}}$ & $\bm{81.61} {\scriptstyle \pm \bm{0.16}}$ &
            $\bm{54.00} {\scriptstyle \pm \bm{0.32}}$ &
            $\bm{53.87} {\scriptstyle \pm \bm{0.67}}$ \\
         \bottomrule
    \end{tabular}
\end{table*}

We consider a setting with $m=100$ clients, two model architectures ResNet-110 and MSDNet, across the \cifar/100 datasets and two heterogeneity levels.
We retain the four groups of heterogeneous clients, which are assigned four different model sizes.
This setup is similar to the one used in \scalefl~\cite{ilhan2023scalefl}.
We then apply the client computational budgets in conjunction with heterogeneous models, to simulate realistic scenarios where other system factors may still induce varying budgets (\eg intermittent bandwidth).
We sample $\cp_i \in [5,15]$ and we set $\cp'_i = 5$.
We report the final accuracy achieved after $1000$ and $2000$ communication rounds for the \cifar and the \cifarH datasets respectively in \Cref{tab:gel_vs_scalefl} and we present their learning curves in \Cref{fig:gel_vs_scalefl}.

From \Cref{tab:gel_vs_scalefl}, we observe that \sys enables \scalefl to achieve nearly $2\%$ higher accuracy points for the \cifar dataset, and about $5\%$ for the \cifarH dataset when using the ResNet-110 architecture.
For both the \scalefl and its combination with \sys, using MSDNet results in higher accuracy than ResNet-110.
Nevertheless, even with MSDNet, \sys brings over $2\%$ improvement in accuracy points to \scalefl under the same number of rounds.
\Cref{fig:gel_vs_scalefl} highlights the accelerated convergence achieved with \sys, showcasing its generality and effectiveness when combined with \scalefl. %
\section{Conclusion}
\label{sec:conclusions}

We designed \sys, our guess and learn algorithm that addresses slow convergence in challenging heterogeneous \FL settings.
The novelty of \sys lies in its gradient-free guessing, thus speeding up convergence \emph{at no cost}, while compensating for low-budget clients.
We demonstrated the wide applicability of \sys by successfully implementing it on top of several state-of-the-art algorithms.
In one of the most promising findings of the paper, we highlighted the utility of \sys as a practical alternative to exhaustive tuning.
Future research directions include exploring the applicability of \sys in other \FL setups, \eg asynchronous FL~\cite{MLSYS2022_f340f1b1,pmlr-v151-nguyen22b} for controlling the staleness of updates.

\section*{Acknowledgments}
This work was supported by funding from the Swiss National Science Foundation under the project ``FRIDAY: Frugal, Privacy-Aware and Practical Decentralized Learning'', SNSF proposal No. 10.001.796.

\bibliography{references}
\bibliographystyle{IEEEtran}

\noindent\begin{minipage}{\linewidth}
\begin{IEEEbiography}[{\includegraphics[width=1in,height=1.1in,clip]{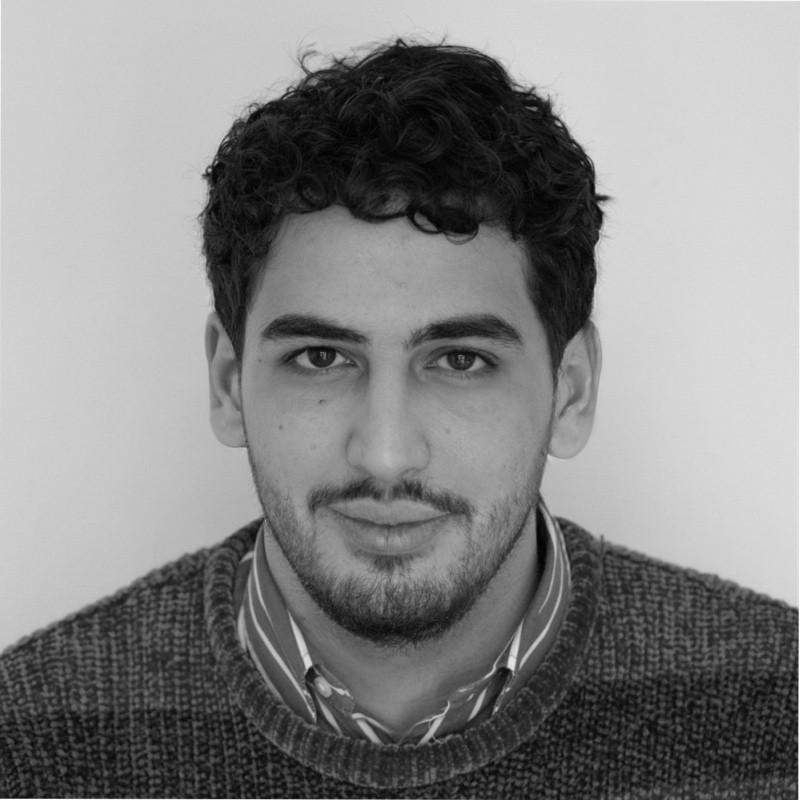}}]{Mohamed Yassine Boukhari}
received his BSc in Computer Science degree from the Swiss Federal Institute of Technology in Lausanne (EPFL),  Switzerland and is currently pursuing the joint MSc degree in Cybersecurity from ETH Zurich and EPFL. His research interests include distributed systems and networking.
\end{IEEEbiography}
\end{minipage}
\noindent\begin{minipage}{\linewidth}
\begin{IEEEbiography}[{\includegraphics[width=1in,height=1.15in,clip,keepaspectratio]{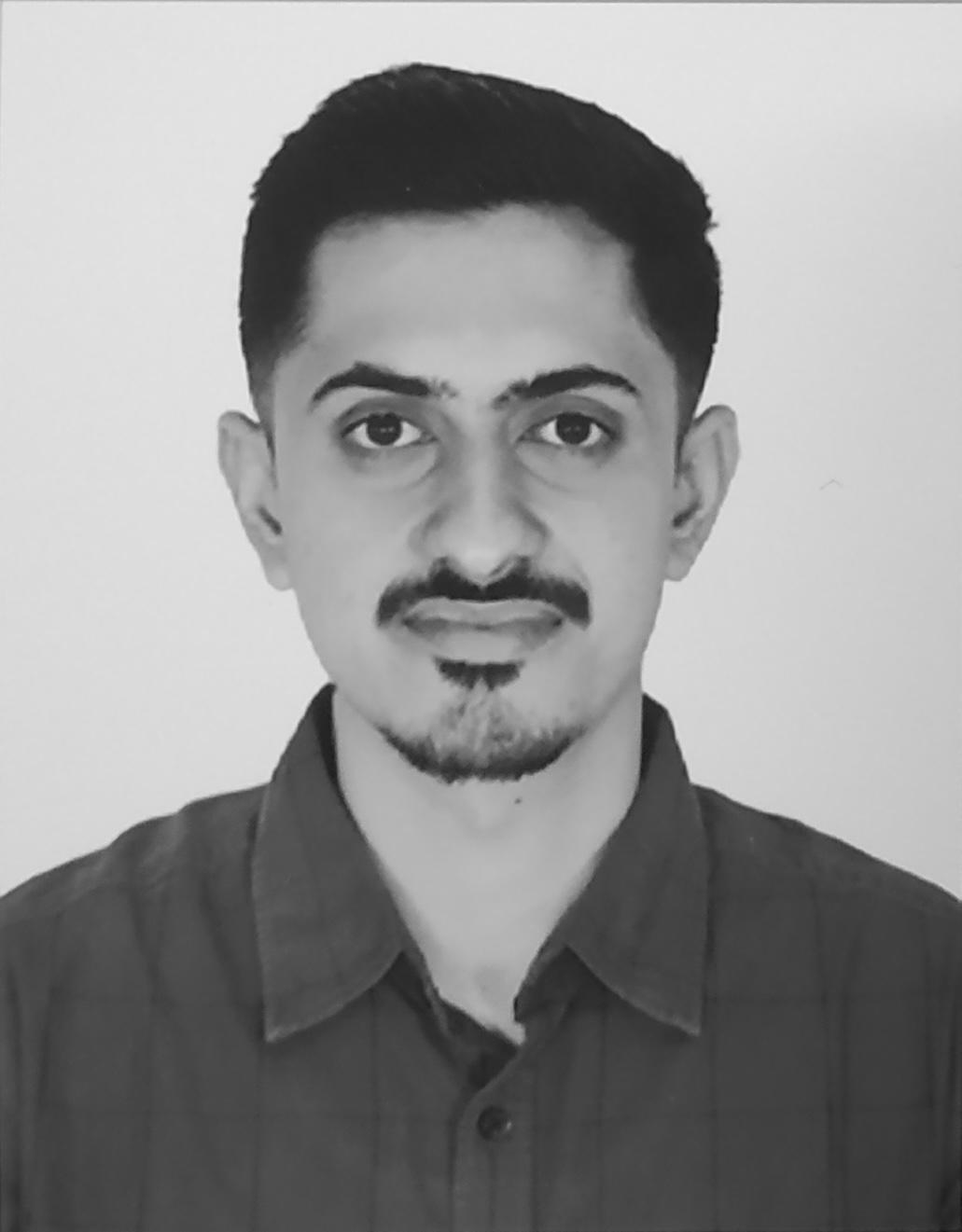}}]{Akash Dhasade}
	received his B.Tech degree in Computer Science and Engineering from the Indian Institute of Technology Tirupati in 2019. He is currently pursuing his PhD at EPFL, Switzerland, supervised by Prof. Dr. Anne-Marie Kermarrec. His research interests include distributed machine learning and systems for machine learning with a focus on federated learning.
\end{IEEEbiography}
\end{minipage}
\noindent\begin{minipage}{\linewidth}
\begin{IEEEbiography}[{\includegraphics[width=1in,height=1.25in,clip,keepaspectratio]{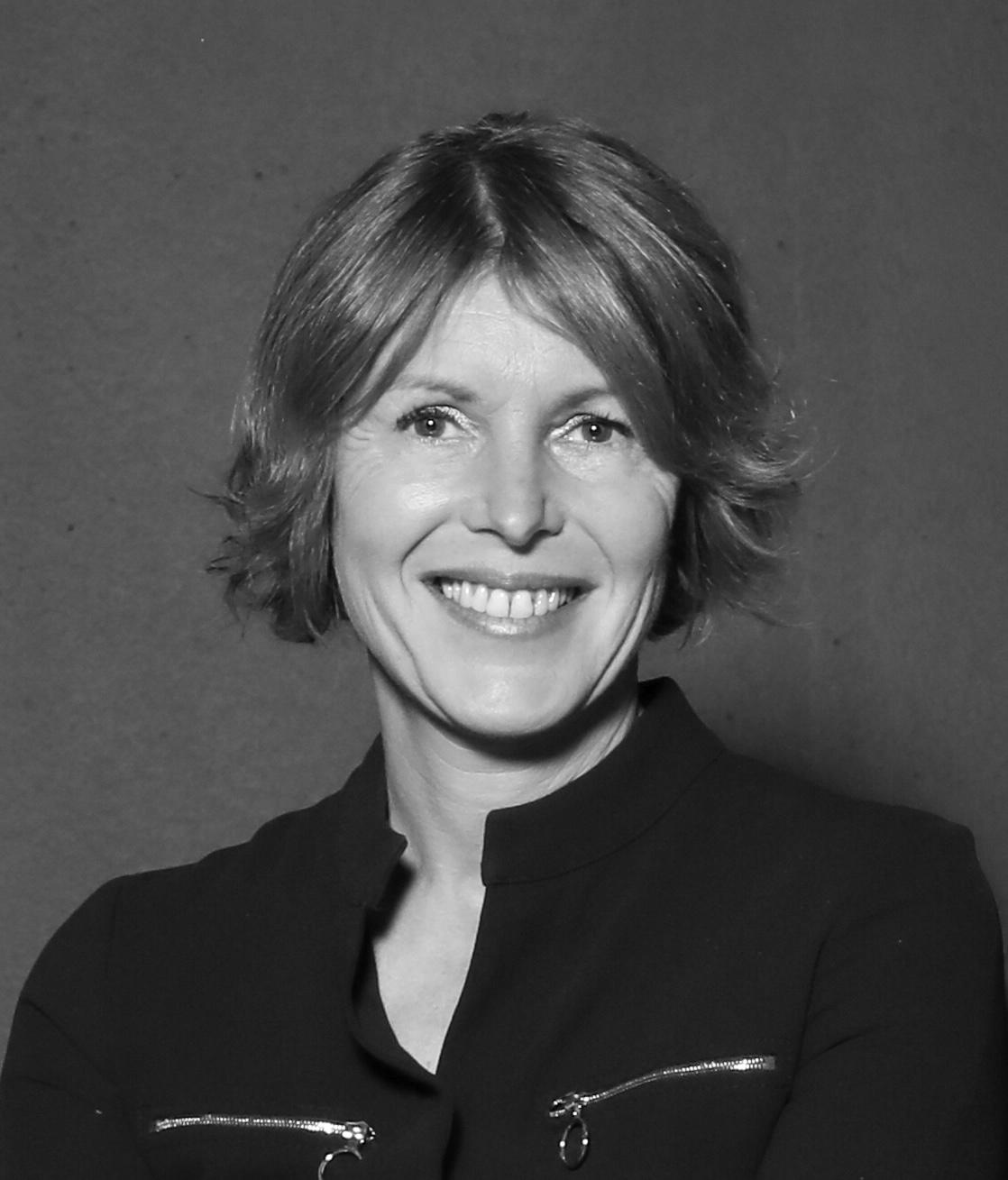}}]{Anne-Marie Kermarrec}
	is Professor at EPFL  since January 2020. Before that she was the CEO of the Mediego startup that she founded in April 2015. Mediego provides content personalization services for online publishers. She was a Research Director at Inria, France from 2004 to 2015. She got a Ph.D. thesis from University of Rennes (France), and has been with Vrije Universiteit, NL and Microsoft Research Cambridge, UK. Anne-Marie received an ERC grant in 2008 and an ERC Proof of Concept in 2013. She received the Montpetit Award in 2011 and the Innovation Award in 2017 from the French Academy of Science. She has been elected to the European Academy in 2013 and named ACM Fellow in 2016. Her research interests are in large-scale distributed systems,  epidemic algorithms,  peer-to-peer networks and system support for machine learning.
\end{IEEEbiography}
\end{minipage}
\noindent\begin{minipage}{\linewidth}
	\begin{IEEEbiography}[{\includegraphics[width=1in,height=1.25in,clip,keepaspectratio]{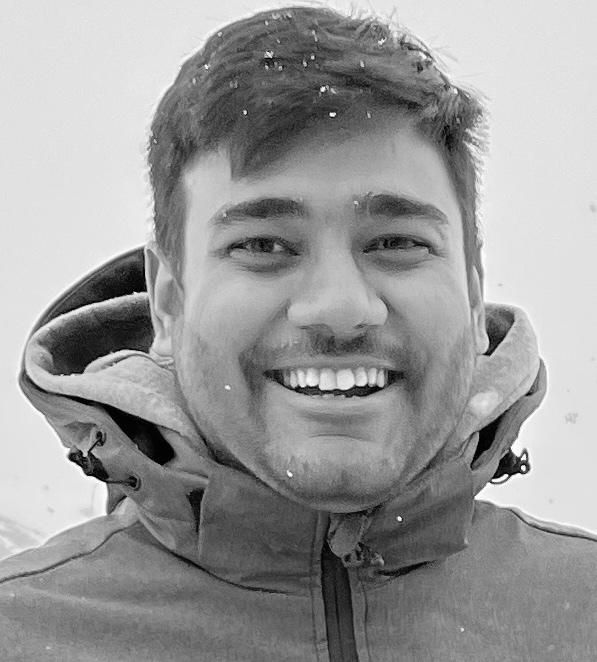}}]{Rishi Sharma}
		received the B.Tech degree from Indian Institute of Technology Mandi in 2021. He is currently pursuing his PhD at EPFL, Switzerland, supervised by Prof. Dr. Anne-Marie Kermarrec. His research interests include parallel and distributed computing, distributed machine learning, and decentralized systems.
	\end{IEEEbiography}
\end{minipage}
\noindent\begin{minipage}{\linewidth}
\begin{IEEEbiography}[{\includegraphics[width=1in,height=1.25in,clip,keepaspectratio]{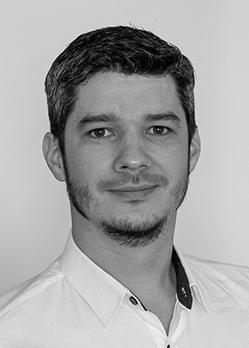}}]{Rafael Pires}
	(Member, IEEE) received the BSc and MSc degrees in Brazil, and the PhD degree in computer science from the University of Neuchâtel, Switzerland. He is a lecturer and researcher with the Swiss Federal Institute of Technology in Lausanne (EPFL), Switzerland. Pires has worked in several aspects of distributed systems, particularly in confidential computing. His current research interests lie in the privacy, scalability and efficiency of systems for machine learning.
\end{IEEEbiography}
\end{minipage}
\noindent\begin{minipage}{\linewidth}
\begin{IEEEbiography}[{\includegraphics[width=1in,height=1.25in,clip,keepaspectratio]{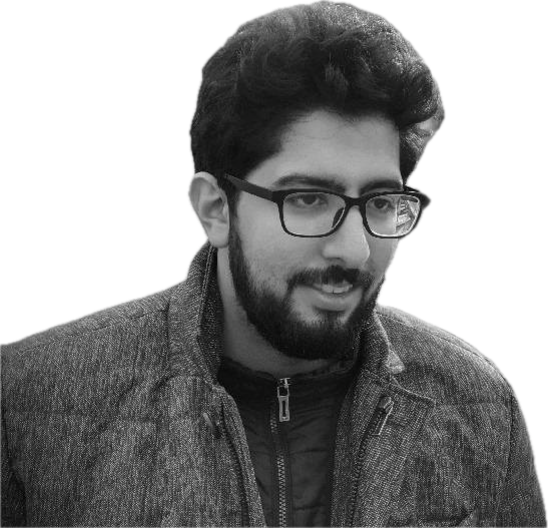}}]{Othmane Safsafi}
	graduated from ENS Paris with a major in Mathematics and minor in Computer Science. He received his PhD in theory of random graphs at Sorbonne University and was a postdoctoral researcher at EPFL. Currently, he works as a quantitative researcher, applying probability and machine learning algorithms to finance.
\end{IEEEbiography}
\end{minipage}

\clearpage
\appendix

\subsection{List of optimizers}
\label{subsec:appendix_opt_list}
\Cref{tab:list_of_algos} provides a list of all \ac{FL} optimization-based approaches with their respective client and server optimizers considered in this work.
\begin{table*}[h]
	\centering
	\caption{List of algorithms referred in this paper and their \gel versions. Vanilla \sgd is referred as just \sgd while \sgdm stands for the \sgd with momentum optimizer. We adopt the acronym name \fedavgCM.  Further, we consider the version of \fednova with the below client and server optimizers due to its superior reported performance~\cite{NEURIPS2020_564127c0}.}
	\label{tab:list_of_algos}
	\begin{tabular} {c  c  c  c}
			\toprule
			\multirow{2}{*}{Algorithm} & \multicolumn{2}{c}{\copt ($\eta_l$)} & \multirow{2}{*}{\sopt ($\eta$)} \\
			\cmidrule(l){2-3}
			& Default & Algorithm + \gel & \\
			\midrule
			\fedavg~\cite{mcmahan2017communication} & \sgd & \sgdm with guessed updates & \sgd \\
			\fedavgCM~\cite{reddi2021adaptive}  & \sgdm &  \sgdm with guessed updates & \sgd \\
			\fedprox~\cite{li2018federated} & Proximal \sgdm & Proximal \sgdm with guessed updates & \sgd \\
			\fednova~\cite{NEURIPS2020_564127c0} & \sgdm & \sgdm with guessed updates & \sgd \\
			\fedyogi~\cite{reddi2021adaptive} & \sgd & \sgdm with guessed updates & \yogi \\
			\bottomrule
		\end{tabular}
\end{table*}

\subsection{Detailed convergence analysis for \fedavgCMG algorithm}
\label{subsec:appendix_proof}

In this section, we detail the convergence result for the \mbox{\fedavgCMG} algorithm presented in \Cref{subsec:convergence_analysis}.
To begin, we first revisit the federated optimization setting.
\subsubsection{The federated optimization setting}
The goal of \FL is to minimize the following objective function with a total of $m$ clients:
\begin{equation}
	\label{eqn:fl_obj}
	\min_{\xsym \in \mathbb{R}^d} \left[F(\xsym) := \sum_{i = 1}^{m} p_i F_i(\xsym)\right]
\end{equation}
where $F_i(\xsym) = \frac{1}{n_i} \sum_{\xi \in \mathcal{D}_i} f_i(\xsym; \xi)$ is the local objective function on the $i^{th}$ client, $p_i = n_i/n$ denotes the relative sample size and $n = \sum_{i=1}^m n_i$. The function $f_i$ represents the loss function (possibly non-convex) on client $i$ defined by the learning model $\xsym$ and samples $\xi$ taken from the local dataset $\mathcal{D}_i$.

Learning occurs in repetitions of communication rounds where in the $t$-th communication round, the server selects a subset $S^{(t)}$of available clients and broadcasts the global model $\xsym^{(t, 0)}$ for local training.
Generally, the number of clients selected $|S^{(t)}| = C$ is kept fixed across communication rounds.
Further, the server requests a fixed computation in a number of steps $\cp$ and waits a stipulated time window to receive updates from the selected clients~\cite{mcmahan2017communication, MLSYS2019_bd686fd6}.
However, as described in \Cref{sec:introduction}, each client manages to perform only a portion of the requested computation, which we denote $\tau_i$.
It corresponds to the computational budget of the $i$-th client, \ie, the number of local learning steps that this client is able to perform in the stipulated time window.
Thus, each client performs a different number of local steps $\{\cp_i\}_{i=1}^C$.

\subsubsection{Heterogeneous federated optimization framework}
\label{subsec:appendix_htr_fed_opt}
Wang \etal~\cite{NEURIPS2020_564127c0} proposed the general theoretical framework to analyze federated algorithms under heterogeneous client budgets.
In this framework, \FL algorithms can be expressed using a general rule as follows:
\begin{equation}
	\label{eqn:general_update}
	\boldsymbol{x}^{(t+1, 0)} = \boldsymbol{x}^{(t, 0)} - \tau_{\text{eff}} \sum_{i = 1}^{m} w_i . \clr \dsym_i^{(t)}
\end{equation}
which optimizes
\begin{equation}
	\label{eqn:surr_obj}
	\widetilde{F}(\xsym) = \sum_{i = 1}^{m} w_i F_i(\xsym)
\end{equation}
where $\dsym_i^{(t)}$ is the normalized gradient, $w_i$'s are aggregation weights and $\cpeff$ is the effective step size.
The normalized gradient is defined as
\begin{equation*}
\dsym_i^{(t)} = \frac{\Gsym_i^{(t)}\asym_i}{||\asym_i||_1}
\end{equation*}
where the matrix $\Gsym_i^{(t)} = [\gsym_i^{(t, 0)}, \gsym_i^{(t, 1)},\ldots,\gsym_i^{(t, \tau_{i-1})}] \in \mathbb{R}^{d \times \tau_i}$ stacks all local stochastic gradients, the vector $\asym_i \in \mathbb{R}^{\tau_i}$ defines the coefficients of these gradients and $||\asym_i||_1$ is the $\ell_1$ norm of the vector $\asym_i$.
Any \FL algorithm whose accumulated local changes $\Delta_i^{(t)} = \xsym_i^{(t, \cp_i)} - \xsym_i^{(t, 0)}$ can be written as a linear combination of local gradients is subsumed by this formulation.

Previous \FL algorithms can be shown to be special cases of this formulation obtained by substituting appropriate values of $w_i$, $\cpeff$, and $\asym_i$.
Specifically, given the value of $\asym_i$, Wang \etal ~\cite{NEURIPS2020_564127c0} show that \FL algorithms take on the following values for $w_i$ and $\cpeff$,
\begin{align*}
	&w_i = \frac{p_i ||\asym_i||_1}{\sum_{i = 1}^m p_i ||\asym_i||_1} \\
	&\cpeff = \sum_{i = 1}^m p_i ||\asym_i||_1.
\end{align*}
Note that the specification of $\asym_i$ defines all variables in the general update rule of equation~\ref{eqn:general_update}.

\subsubsection{Proof of Lemma \ref{lem:linear_comb}: Accumulated updates in \fedavgCMG algorithm}
\label{subsec:proof_of_lemma}

Now we prove that the update rule for the \fedavgCMG algorithm forms a linear combination of gradients, allowing us to apply the above general theoretical framework.
To begin, recall the \sgd with momentum equation~\ref{eqn:mom_on_clients} and equation~\ref{eqn:mom_model_update_on_clients}:

\begin{equation*}
	\begin{split}
		\vsym_i^{(t, k+1)} &= \alpha \vsym_i^{(t, k)} - \clr \gsym_i^{(t, k)} \\
		\xsym_i^{(t, k+1)} &= \xsym_i^{(t, k)} + \vsym_i^{(t, k+1)}
	\end{split}
\end{equation*}
By a simple recursion, we get:
\begin{equation*}
	\begin{split}
		\vsym_i^{(t, k)} &= \alpha \vsym_i^{(t, k-1)} - \clr \gsym_i^{(t, k-1)} \\
		&= -\clr \sum_{j=0}^{k-1}\alpha^{k-1-j}\gsym_i^{(t, j)} \hspace{5mm}(\text{where } \vsym_i^{(t, 0)}=0)
	\end{split}
\end{equation*}
Hence:
\begin{equation*}
	\begin{split}
		\xsym_i^{(t, \tau_i)} -  \xsym_i^{(t, 0)} &= \sum_{k=1}^{\tau_i}\vsym_i^{(t, k)} \\
		&= -\clr\sum_{k=1}^{\tau_i}\sum_{j=0}^{k-1}\alpha^{k-1-j}\gsym_i^{(t, j)}
	\end{split}
\end{equation*}
Moreover, recall from equation~\ref{eqn:gel_final} that:
\begin{equation*}
	\begin{split}
		\xsym_i^{(t, \cp_i+\cp'_i)} &- \xsym_i^{(t, \cp_i)} = \left(\alpha \frac{1-\alpha^{\cp'_i}}{1-\alpha} \right) \vsym_i^{(t, \cp_i)} \\
		&= -\clr \left(\alpha \frac{1-\alpha^{\cp'_i}}{1-\alpha} \right) \sum_{j=0}^{\cp_i-1}\alpha^{\cp_i-1-j}\gsym_i^{(t, j)}
	\end{split}
\end{equation*}
By combining the two equations above, we have:
\begin{equation*}
	\begin{split}
		&\xsym_i^{(t, \cp_i+\cp'_i)} - \xsym_i^{(t, 0)} =  \\
		& -\clr \left(\alpha \frac{1-\alpha^{\cp'_i}}{1-\alpha} \right) \sum_{j=0}^{\cp_i-1}\alpha^{\cp_i-1-j}\gsym_i^{(t, j)}  \\
		&-\clr\sum_{k=1}^{\tau_i}\sum_{j=0}^{k-1}\alpha^{k-1-j}\gsym_i^{(t, j)}
	\end{split}
\end{equation*}
Rewriting the second term on the right-hand side:
\begin{equation*}
	\begin{split}
		&\xsym_i^{(t, \cp_i+\cp'_i)} - \xsym_i^{(t, 0)} =  \\
		& -\clr \left(\alpha \frac{1-\alpha^{\cp'_i}}{1-\alpha} \right) \sum_{j=0}^{\cp_i-1}\alpha^{\cp_i-1-j}\gsym_i^{(t, j)}  \\
		&-\clr\sum_{j=0}^{\tau_i-1}g_i^{(t, j)} \sum_{k=0}^{\cp_i-j-1}\alpha^{k}
	\end{split}
\end{equation*}
Putting the coefficients together:
\begin{equation*}
	\begin{split}
		&\xsym_i^{(t, \cp_i+\cp'_i)} - \xsym_i^{(t, 0)} =  \\
		&-\clr\sum_{j=0}^{\tau_i-1} \left[\left(\alpha \frac{1-\alpha^{\cp'_i}}{1-\alpha} \right) \alpha^{\cp_i-1-j} + \left(\sum_{k=0}^{\cp_i-j-1}\alpha^{k}\right)\right] \gsym_i^{(t, j)} \\
		&=-\clr\sum_{j=0}^{\tau_i-1} \left[\left(\alpha \frac{1-\alpha^{\cp'_i}}{1-\alpha} \right) \alpha^{\cp_i-1-j} + \frac{1-\alpha^{(\cp_i-j)}}{1-\alpha}\right] \gsym_i^{(t, j)} \\
		&=-\clr\sum_{j=0}^{\tau_i-1} \left[\frac{\alpha^{\cp_i-j} - \alpha^{\cp'_i + \cp_i - j} +1 - \alpha^{\cp_i - j}}{1-\alpha}\right] \gsym_i^{(t, j)} \\
		&=-\clr\sum_{j=0}^{\tau_i-1} \left[\frac{1 -  \alpha^{\cp'_i + \cp_i - j}}{1-\alpha}\right] \gsym_i^{(t, j)}
	\end{split}
\end{equation*}
Thus, the update rule of \fedavgCMG can be expressed as a linear combination of gradients where the coefficient of $\gsym_i^{(t, j)}$ is $\frac{1-\alpha^{\cp'_i+\cp_i-j}}{1-\alpha}$.
With this, we obtain the coefficient vector $\asym_i$:
\begin{equation}
	\label{eqn:appendix_final_coeff}
	\asym_i = [1-\alpha^{\cp'_i+\cp_i}, 1-\alpha^{\cp'_i+\cp_i-1}, ..., 1-\alpha^{\cp'_i+1}]/(1-\alpha)
\end{equation}
This proves our Lemma~\ref{lem:linear_comb} presented in \Cref{subsec:convergence_analysis}.
Thus, \fedavgCMG can also be expressed using the update rule \Cref{eqn:general_update} which leads us to the following result.

\subsubsection{Final convergence result}

\cite{NEURIPS2020_564127c0} show that for \FL algorithms whose update rule follows \Cref{eqn:general_update}, thus subsuming \mbox{\fedavgCMG}, the following convergence result holds under standard assumptions in the federated optimization literature~\cite{bottou2018optimization, NEURIPS2020_564127c0,karimireddy2021mime}.

\begin{assumption}[Smoothness]\label{assump:smooth}
	$||\nabla F_i(\xsym) - \nabla F_i(\ysym)|| \leq L ||\xsym-\ysym||, \forall i \in \{1,2,\ldots,m\}$.
\end{assumption}

\begin{assumption}[Unbiased gradients and bounded variance]\label{assump:var}
	$\mathbb{E}_\xi[g_i(\xsym|\xi)] = \nabla F_i(\xsym)$ and $\mathbb{E}_\xi[||g_i(\xsym|\xi) - \nabla F_i(\xsym)||^2] \leq \sigma^2, \forall i \in \{1,2,\ldots,m\}, \sigma^2 \geq 0$
\end{assumption}

\begin{assumption}[Bounded Dissimilarity]\label{assump:dissimilarity}
	For any set of weights $\{w_i\geq 0\}_{i=1}^m, \sum_{i=1}^m w_i =1$, there exist constants $\bnda \geq 1, \bndb \geq 0$ such that $\sum_{i=1}^\nworkers w_i ||\tg_i(\xsym)||^2 \leq \bnda||\sum_{i=1}^\nworkers w_i\tg_i(\xsym)||^2 + \bndb$
\end{assumption}

\begin{theorem}[\textbf{Convergence to the $\widetilde{F}(\x)$'s Stationary Point}]\label{thm:general}
	Under Assumptions \ref{assump:smooth} to \ref{assump:dissimilarity}, any federated optimization algorithm that follows the update rule (\ref{eqn:general_update}), will converge to a stationary point of a surrogate objective $\surloss(\x)=\sum_{i=1}^\nworkers w_i \obj_i(\x)$. More specifically, if the total communication rounds $T$ is pre-determined and the learning rate $\clr$ is small enough $\clr = \sqrt{\nworkers/\cpavg T}$ where $\cpavg=\frac{1}{\nworkers}\sum_{i=1}^\nworkers \cp_i$, then the optimization error will be bounded as follows:
	\begin{align}
		&\min_{t\in[T]} \mathbb{E}\| \nabla \surloss(\x^{(t,0)}) \|^2 \leq \mathcal{O} \left(\frac{\cpavg/\teff}{\sqrt{\nworkers \cpavg T}}\right) +\nonumber\\
		&\mathcal{O}\left(\frac{A\vbnd}{\sqrt{\nworkers \cpavg T}} \right) + \mathcal{O}\left(\frac{\nworkers B\vbnd}{\cpavg T}\right) + \mathcal{O}\left(\frac{\nworkers C \bndb}{\cpavg T}\right)
		\label{eqn:general_error}
	\end{align}
	where $\mathcal{O}$ swallows all constants (including $\lip$), and quantities $A, B, C$ are defined as follows:
	\begin{align}
		&A = \nworkers\cpeff\sum_{i=1}^\nworkers \frac{w_i^2 ||\asym_i||_2^2}{||\asym_i||_1^2},
		B = \sum_{i=1}^\nworkers w_i (||\asym_i||_2^2 - a_{i,-1}^2), \nonumber\\
		&C = \max_i\{||\asym_i||_1^2-||\asym_i||_1 a_{i,-1} \}
	\end{align}
\end{theorem}

Thus, it follows that \fedavgCMG also converges at an asymptotic rate of $\mathcal{O}(1/\sqrt{m \bar{\cp} T})$ where we substitute $\asym_i$ from our derivation in \Cref{eqn:appendix_final_coeff}.
We also note that the algorithm converges to a surrogate objective $\widetilde{F}(\xsym)$ (equation~\ref{eqn:surr_obj}) over the true objective (equation~\ref{eqn:fl_obj}).
The mismatch between objectives arises from heterogeneous $\tau_i$ and is not due to \sys.
Traditional algorithms like \fedavg and \fedprox also face this inconsistency in the scenario of heterogeneous steps.
This was one of the critical findings of the general theoretical framework presented in \Cref{subsec:appendix_htr_fed_opt}.
However, our empirical results in \Cref{sec:exp_res} indicate a modest impact of this inconsistency as all algorithms manage to converge to a similar accuracy after appropriate tuning.
Additionally, one way to get exact convergence is to use \sys on top of \fednova~\cite{NEURIPS2020_564127c0}, which eliminates the objective inconsistency.
Results for \sys combined with \fednova are also presented in \Cref{subsec:gel_against_sota_eval}

\subsection{Hyperparameter tuning}
\label{subsec:appendix_hparams}

\subsubsection{Fixed parameters}
The number of selected clients $(C)$ per round is fixed at 20 in all experiments, except for our experiments involving the \agnews dataset and \scalefl, where we use $C=10$.
Batch sizes of $5$, $20$, $20$, $16$ and $8$ are used for the Synthetic, FEMNIST, Shakespeare, \cifar/100 and \agnews datasets, respectively.
In all instances of the \sgdm optimizer, the momentum parameter is set to $0.9$.
Similarly, we let $\beta_1 = 0.9$ and $\beta_2 = 0.99$ for the \yogi optimizer.
Lastly, we fix the adaptivity parameter to $10^{-3}$ since it was shown to perform nearly as well as other values~\cite{reddi2021adaptive}, saving significant tuning effort.

\subsubsection{Tuning learning rate}
We tune the client learning rate $(\clr)$ for the \fedavgCM algorithms for all datasets and tune the \fedavg separately for datasets in the LEAF benchmark for \Cref{fig:vanilla_sgd}.
We tried several values to obtain the following final search space, where Table~\ref{tab:regular_best_learning_rates} lists the best $\clr$.

\noindent
\textbf{FEMNIST}:
\[\eta_l \in \{0.001, 0.005, 0.01, 0.02, 0.03, 0.06\}\]
\textbf{Synthetic}:
\[\eta_l \in \{0.001, 0.005, 0.01, 0.02, 0.05, 0.08, 0.1\}\]
\textbf{Shakespeare}:
\[\eta_l \in \{0.01, 0.05, 0.1, 0.3, 0.5, 0.6, 0.8\}\]
\textbf{\cifar/100}:
\[
\eta_l \in \{10^{-1}, 10^{-2}, 10^{-3}, 10^{-4}\}
\]
\textbf{\agnews}:
\[\eta_l \in \{10^{-3}, 10^{-4}, 10^{-5}\}\]

\begin{table}[h]
	\centering
	\caption{The best learning rate obtained after hyperparameter tuning for the \fedavg and \fedavgCM algorithm on the different datasets.}
	\label{tab:regular_best_learning_rates}
	\begin{tabular}[t]{l*{3}{c}}
		\toprule
		Dataset & \fedavg & \fedavgCM \\
		& $\clr$ & $\clr$ \\
		\midrule
		FEMNIST     & 0.06 & 0.02 \\
		Synthetic   &  0.1 & 0.01 \\
		Shakespeare    & 0.8 & 0.3 \\
		\cifar/100 & -- & 0.01 \\
		\agnews & -- & 0.001  \\
		\bottomrule
	\end{tabular}
\end{table}

\textbf{\fedopt framework.} For the experiments using the \fedopt framework (\Cref{subsec:eval_gel_in_fedopt}), we tune both the client $(\clr)$ and the server learning rate $(\slr)$ for the \fedavgCM and \fedyogi algorithms.
Similar to~\cite{reddi2021adaptive}, we select the best parameters as the ones that minimize the average training loss over the last 100 rounds of training.
We run 1000 rounds of training on the FEMNIST dataset over the following grid:
\[\eta_l \in \{10^{-3}, 10^{-2.5}, \dots, 10^{0.5}\}\]
\[\eta \in \{10^{-3}, 10^{-2.5}, \dots, 10^{1}\}\]

We chart the test accuracy obtained on this grid in Figure~\ref{fig:gelimproveshparam}.
We report the best values obtained in Table~\ref{tab:fedopt_best_learning_rates} and use these values in our experiments of \Cref{subsec:eval_gel_in_fedopt}.

\begin{table}[h]
	\centering
		\caption{The base-10 logarithm of the client ($\eta_l$) and server ($\eta$) learning rate combinations after tuning. }
	\label{tab:fedopt_best_learning_rates}
	\begin{tabular}[t]{l*{5}{c}}
		\toprule
		Dataset & \multicolumn{2}{c}{\fedavgCM} & \multicolumn{2}{c}{\fedyogi} \\
		& $\clr$ & $\slr$& $\clr$ & $\slr$ \\
		\midrule
		FEMNIST    &  -\thlf &  0 & -\thlf & -2 \\
		\bottomrule
	\end{tabular}
\end{table}

\subsubsection{\fedprox proximal parameter $\mu$}
The proximal term $\mu$ restricts the trajectory of the local updates by constraining them to be closer to the global model, thus a large value of $\mu$ can slow down convergence by forcing the updates to stay close to the starting point.
In our settings, clients do not perform excessive local steps (restricted by computational budgets) and hence, the client models will not drift far away from the server model.
We set $\mu$ to a fixed value of $0.01$ from the limited set of candidates $\{0.001, 0.01, 0.1, 1.0\}$ used in previous works~\cite{li2018federated}.

\end{document}